\documentclass{article}
\usepackage{iclr2027_conference,times}

\usepackage{amsmath,amsfonts,bm}

\def\eqref#1{equation~\ref{#1}}

\def\1{\bm{1}}

\DeclareMathAlphabet{\mathsfit}{\encodingdefault}{\sfdefault}{m}{sl}
\SetMathAlphabet{\mathsfit}{bold}{\encodingdefault}{\sfdefault}{bx}{n}

\usepackage{graphicx}
\usepackage{wrapstuff}
\usepackage{booktabs}
\usepackage{colortbl}
\usepackage{capt-of}
\usepackage{multirow}
\usepackage{enumitem}
\usepackage{hyperref}
\usepackage{url}

\newcount\paperversion
\title{What Matters for World Modeling in Vision-Language-Action Models?}

\author{%
  \makebox[\dimexpr\textwidth-2\tabcolsep\relax][c]{%
    \begin{tabular}{c}
      Qisen Ma\textsuperscript{\normalfont 1,2}\quad
      Yixiang Chen\textsuperscript{\normalfont 1,2}\quad
      Yuan Xu\textsuperscript{\normalfont 1,2}\quad
      Peiyan Li\textsuperscript{\normalfont 1,2}\quad
      Jiabing Yang\textsuperscript{\normalfont 1,2}\\[2pt]
      Kai Wang\textsuperscript{\normalfont 1,2}\quad
      Jianhua Yang\textsuperscript{\normalfont 1,2}\quad
      Jianlou Si\textsuperscript{\normalfont 3}\quad
      Jun Huang\textsuperscript{\normalfont 3}\\[2pt]
      Jing Liu\textsuperscript{\normalfont 4}\quad
      Nianfeng Liu\textsuperscript{\normalfont 4}\quad
      Yan Huang\textsuperscript{\normalfont 1,2,4}\quad
      Liang Wang\textsuperscript{\normalfont 1,2}
    \end{tabular}%
  }\\[6pt]
  \normalfont\small
  \makebox[\dimexpr\textwidth-2\tabcolsep\relax][c]{%
    \begin{tabular}{c}
      \textsuperscript{1}New Laboratory of Pattern Recognition (NLPR),\\
      Institute of Automation,
      Chinese Academy of Sciences\\
      \textsuperscript{2}School of Artificial Intelligence,
      University of Chinese Academy of Sciences\\[2pt]
      \textsuperscript{3}Alibaba Group \quad
      \textsuperscript{4}FiveAges
    \end{tabular}%
  }%
}

\ifnum\paperversion>0
  \iclrfinalcopy
\fi
\begin{document}

\maketitle
\ifnum\paperversion=1
  \lhead{}
  \renewcommand{\headrulewidth}{0pt}
\fi

\vspace{-10pt}

\begin{abstract}
Integrating world models into Vision-Language-Action models (VLAs) has emerged as a promising direction, enabling policies to anticipate future dynamics beyond reactive observation-to-action mapping.
However, existing approaches vary simultaneously in their world representations, integration paradigms, and policy backbones, making it difficult to determine what truly matters for effective world model integration.
To disentangle these factors, we categorize existing efforts into four representative paradigms, namely Implicit Future Alignment, Explicit Future Query, MoT Future Expert, and External World Predictor.
We instantiate each within a unified framework and systematically evaluate these paradigms with diverse world representations and policy backbones across in-domain, out-of-domain, long-horizon, dynamic, few-shot, and real-world settings.
Our study yields four main findings:
1) world representation effectiveness is paradigm-dependent, whereas 2D semantic representations offer robust, cross-paradigm gains;
2) no single paradigm is universally optimal, yet the decoupled MoT Future Expert offers competitive overall performance, particularly in OOD robustness;
3) world modeling consistently benefits policies built on standard VLMs, while decoupled designs exhibit greater compatibility with pretrained VLAs;
and 4) world modeling improves performance over action-only baselines across multiple settings, with notable gains on long-horizon and dynamic tasks.
Project page and video demonstrations are available at \url{https://dylanmqs.github.io/WM-in-VLAs/}.

\end{abstract}

\section{Introduction}
\label{sec:introduction}

Vision-Language-Action models (VLAs)~\citep{zitkovich2023rt, kim2024openvla, intelligence2025pi} have achieved notable success in robotic manipulation by adapting pretrained Vision-Language Models (VLMs) to continuous physical control.
However, standard VLAs mainly predict actions from the current observation.
Without modeling how the scene evolves under robot interaction, they struggle in long-horizon tasks and fail to adapt to distribution shifts or dynamic environments~\citep{yang2026mantis, chen2026dial, lyu2026lda}.

To overcome this limitation, a promising direction is to integrate world models into VLAs to anticipate world dynamics~\citep{cen2025worldvla,zheng2025flare}.
By incorporating world dynamics, such models have demonstrated improved long-horizon execution~\citep{chen2026lawam}, dynamic manipulation~\citep{fang2026towards}, environmental robustness~\citep{sun2026vlajepa}, and sample efficiency~\citep{chen2026dial}.
Despite these empirical successes, existing studies adopt heterogeneous world representations, integration paradigms, and policy backbones.
Specifically, they combine representations ranging from pixel details to semantic latents with designs such as implicit alignment and explicit conditioning across different VLA backbones~\citep{zheng2025flare, lv2025f1, su2026world, sun2026vlajepa}.
Consequently, the true sources of empirical gains remain obscured, leaving it unclear which design choices fundamentally matter when integrating world modeling into VLAs.

Among these challenges, four core questions stand out:
(1)~\textbf{World Representation}: Is there a universally optimal world representation for robot policy learning, or is representation effectiveness fundamentally coupled with the integration paradigm?
(2)~\textbf{Integration Paradigm}: What are the performance trade-offs among different integration paradigms, and is coupled or decoupled world modeling more effective?
(3)~\textbf{Backbone Generalization}: Do world modeling integration paradigms remain effective across policy backbones, spanning models without robot pretraining and pretrained foundation VLAs?
(4)~\textbf{Comprehensive Performance}: How do different world modeling integration paradigms perform under diverse task conditions?

\begin{figure*}[t]
    \centering
    \includegraphics[width=1.0\textwidth]{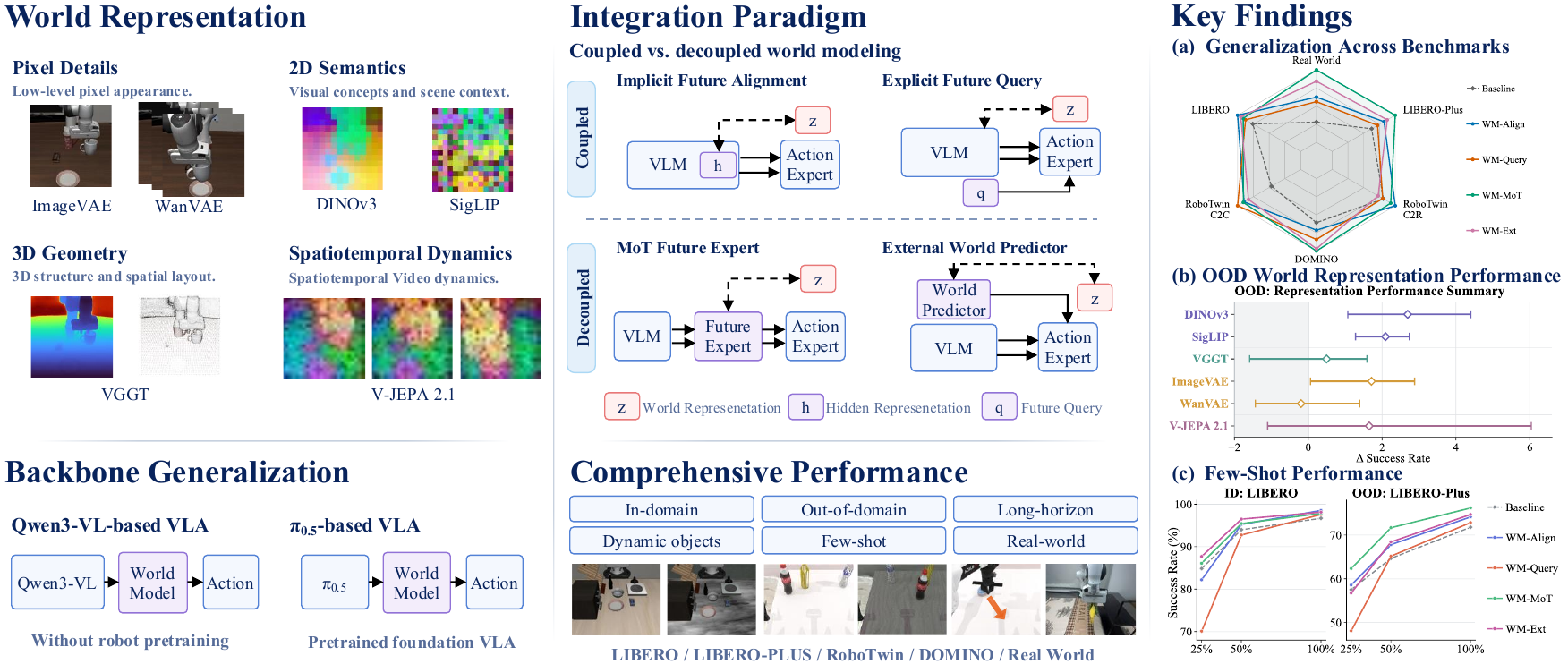}
    \caption{Systematic empirical investigation of world modeling in vision-language-action policies across four core dimensions, spanning world representation, integration paradigms, backbone generalization, and comprehensive performance across diverse settings.}
    \label{fig:vlaforesight_teaser}
    \vspace{-20pt}
\end{figure*}

In this work, we present the first systematic, controlled empirical investigation of world modeling in VLA policies, as shown in Figure~\ref{fig:vlaforesight_teaser}.
To isolate world modeling choices, we conduct our study under a unified pipeline with a shared policy backbone and training protocol within each comparison, systematically addressing the four core questions:
First, to answer World Representation, we examine diverse world representations spanning pixel details (ImageVAE~\citep{tian2024visual}, WanVAE~\citep{wan2025wan}), 2D semantics (DINOv3~\citep{simeoni2025dinov3}, SigLIP~\citep{zhai2023sigmoid}), 3D spatial geometry (VGGT~\citep{wang2025vggt}), and spatiotemporal video dynamics (V-JEPA~2.1~\citep{mur2026vjepa21}).
Second, to answer Integration Paradigm, we formalize existing architectures into four integration paradigms defined by how world dynamics are conditioned on and decoupled from the policy, namely Implicit Future Alignment, Explicit Future Query, MoT Future Expert, and External World Predictor.
We instantiate these paradigms as representative designs within our unified framework.
Third, to answer Backbone Generalization, we benchmark world modeling paradigms across representative policy backbones spanning a standard VLM backbone (Qwen3-VL-2B~\citep{bai2025qwen3vl}) and a dedicated robot foundation VLA ($\pi_{0.5}$~\citep{intelligence2025pi}).
Finally, to answer Comprehensive Performance, we evaluate these integration paradigms across diverse settings spanning in-domain and out-of-domain distributions, long horizons, dynamic objects, few-shot data, as well as real-world experiments.

Our extensive experiments yield four primary findings corresponding to our core questions:
(1)~\textbf{World Representation}:
No single world representation is universally optimal.
Pixel-level features do not consistently translate in-domain gains into OOD robustness, while geometric and spatiotemporal representations are sensitive to the integration paradigm. DINOv3 and SigLIP provide consistent gains across paradigms.
(2)~\textbf{Integration Paradigm}:
No paradigm dominates across all settings, yet the decoupled MoT Future Expert offers competitive overall performance, particularly in OOD robustness.
Implicit Future Alignment is lightweight but less effective on dynamic tasks, while Explicit Future Query is more task-dependent.
The External World Predictor achieves the best low-data in-domain performance.
(3)~\textbf{Backbone Generalization}:
Integration effectiveness differs across policy backbones.
World modeling generally benefits policies built on standard VLMs, whereas decoupled designs offer substantially better compatibility when adapting pretrained foundation VLAs.
(4)~\textbf{Comprehensive Performance}:
World modeling provides clear gains over action-only policies, particularly on long-horizon and dynamic tasks.
We believe these systematic findings provide actionable design principles for integrating world modeling into VLA systems.

\section{Related Work}

\noindent
\textbf{Vision-Language-Action Models.}
Vision-Language-Action models (VLAs)~\citep{kim2025fine, intelligence2025pi, ye2026starvla} build on pretrained Vision-Language Models (VLMs) to predict robotic actions from a language instruction and the current observation.
These models have shifted from autoregressive discrete token decoding~\citep{zitkovich2023rt, kim2024openvla} to continuous diffusion- or flow-based action heads~\citep{liu2024rdt, black2024pi0, intelligence2025pi}.
To enhance policy capabilities, parallel efforts inject 3D priors~\citep{zhen20243d, qu2025spatialvla, li2025bridgevla,li2025spatial}, adopt dual-system architectures~\citep{bjorck2025gr00t, chen2025fast, cui2025openhelix}, insert chain-of-thought reasoning before action generation~\citep{zhao2025cot, huang2026fast, zhong2026acot, wang2026vla}, incorporate latent actions~\citep{bu2025univla, lin2026pixels}, or train on heterogeneous data mixtures~\citep{bu2025univla, zheng2025x, zheng2026egoscale, luo2026joint}.
Despite these advances, standard VLAs inherit the static image--text pretraining of VLMs, lacking the temporal and physical dynamics priors essential for complex manipulation.
This motivates recent efforts to integrate world modeling into policy learning.

\noindent
\textbf{World Modeling in VLAs.}
To overcome the limitations of reactive policies, recent works integrate world models to provide VLAs with physical dynamics.
Depending on how world dynamics are conditioned on and decoupled from the policy, existing approaches can be organized into four paradigms:
\textbf{(1) Implicit Future Alignment} uses world representations as auxiliary supervision to refine the VLM backbone without feeding predictions directly to the action head~\citep{su2026world, xu2026futurevla, sun2026vlajepa, lin2026jepawam}.
\textbf{(2) Explicit Future Query} introduces dedicated query tokens into the VLM to predict future latents, serving as explicit context for action generation~\citep{zhang2025dreamvla, lin2025hif, liu2026last, chen2026dial, luo2026being, ma2026internvla}.
\textbf{(3) MoT Future Expert} co-optimizes parallel action and future experts within a Mixture-of-Transformers architecture, conditioning action generation on the future expert's intermediate hidden states~\citep{lv2025f1, zhang2025unijepa,bi2025motus, won2025dual, ma2026internvla, hu2026bagelvla}.
\textbf{(4) External World Predictor} decouples prediction by employing a standalone world predictor, supplying predicted future states as visual subgoals for the policy~\citep{chen2026lawam}.
However, the field still lacks a systematic and controlled comparison of these paradigms, leaving it unclear how different paradigms choices affect the effectiveness of world modeling in VLAs.

\section{Method}
\label{sec:method}

\subsection{Preliminaries}
\label{subsec:preliminaries}

Standard VLAs typically combine a pretrained VLM backbone with a trainable action expert to map observations and instructions into robot actions.
Given an input observation $o_t \in \mathbb{R}^{H \times W \times 3}$ and an instruction $c$, the backbone embeds them into visual and language tokens, $x^V_{1:N}$ and $x^L_{1:M}$, respectively.
These tokens are processed across successive VLM attention layers to yield joint multimodal representations, from which the action expert decodes a continuous action chunk $\hat{a}_{t:t+K-1}$.

Based on this formulation, we establish a primary baseline by pairing a Qwen3-VL-2B~\citep{bai2025qwen3vl} backbone with a layer-wise cross-DiT action expert, while $\pi_{0.5}$~\citep{intelligence2025pi} is additionally used to evaluate backbone generalization.
The action expert first projects a noisy action chunk into $K$ action tokens $x^A_{1:K}$, which cross-attend to the visual and language tokens at each layer:
\begin{equation}
  \widetilde{x}_{1:K}^A = \operatorname{CrossAttn}\!\left(
  x_{1:K}^A, [x_{1:N}^V; x_{1:M}^L]\right).
  \label{eq:layerwise-cross-attn}
\end{equation}
Following \citet{black2024pi0}, the VLM backbone and action expert are jointly trained using the flow-matching objective $\mathcal{L}_{\mathrm{action}}$.

The following foresight strategies share this unified baseline, differing in how they integrate world modeling into action generation.
To provide a unified supervision target across all strategies, a frozen encoder $E$ extracts the world representation $z_{t:t+T}$ from future observations $o_{t:t+T}$:
\begin{equation}
  z_{t:t+T} = E(o_{t:t+T}).
  \label{eq:future-target}
\end{equation}
All encoders receive the same future frames, ensuring identical temporal coverage.
For image encoders, $E$ processes sampled frames independently, whereas for video encoders, it encodes the sequence jointly.
Each strategy produces a predicted world representation $\hat{z}_{t:t+T}$, trained to maximize its alignment with the target $z_{t:t+T}$ via a cosine similarity objective:
\begin{equation}
  \mathcal{L}_{\text{world}} = 1 - \cos\!\left(\hat{z}_{t:t+T}, z_{t:t+T}\right),
  \label{eq:world-loss}
\end{equation}
The entire framework is optimized with a joint objective:
\begin{equation}
  \mathcal{L} = \mathcal{L}_{\text{action}} + \lambda_{\text{world}} \mathcal{L}_{\text{world}}.
  \label{eq:joint-objective}
\end{equation}

\begin{figure*}[t]
    \centering
    \includegraphics[width=0.9\textwidth]{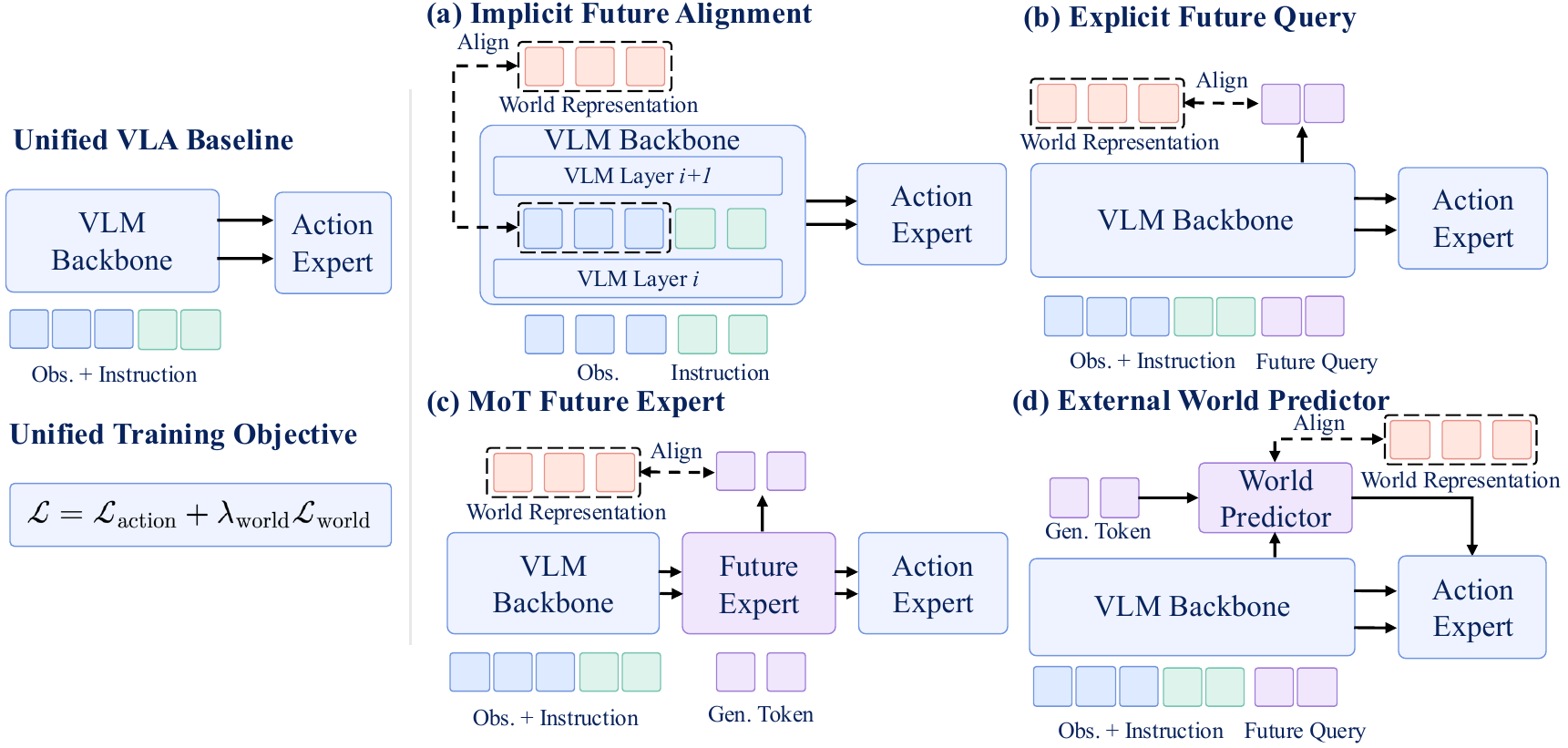}
    \caption{Overview of the four representative world modeling integration paradigms for vision-language-action models: (a)~Implicit Future Alignment, (b)~Explicit Future Query, (c)~MoT Future Expert, and (d)~External World Predictor, categorized by how world dynamics are conditioned on and decoupled from the policy.}
    \label{fig:pic_method}
\end{figure*}

\subsection{Strategy 1: Implicit Future Alignment}
Implicit Future Alignment introduces no dedicated architectural modules, but instead formulates world modeling as an auxiliary representation learning objective that infuses world dynamics directly into the VLM's hidden representations~\citep{su2026world, xu2026futurevla, sun2026vlajepa, lin2026jepawam}, as illustrated in Fig.~\ref{fig:pic_method}(a).

Specifically, this strategy supervises the VLM's visual representations to capture predictive world dynamics.
Formally, given visual tokens $x_{1:N}^V$ from an intermediate layer, a lightweight MLP predictor projects them to align with the target world representation $z_{t:t+T}$:
\begin{equation}
  \mathcal{L}_{\text{world}} = 1 - \cos\!\left( \operatorname{MLP}(x_{1:N}^V), z_{t:t+T} \right).
  \label{eq:implicit-patch-prediction}
\end{equation}
Under this predictive supervision, the intermediate visual representations absorb future world dynamics while preserving current semantic context.
Through the framework's standard cross-attention (Eq.~\ref{eq:layerwise-cross-attn}), action tokens $x_{1:K}^A$ naturally inherit physical foresight by attending to dynamics-enriched visual representations.
At inference, the auxiliary predictor is discarded, leaving the policy deployment identical to the baseline.

\subsection{Strategy 2: Explicit Future Query}
Explicit Future Query performs world modeling through a compact set of learnable future query tokens $x_{1:Q}^Q$ appended to the input sequence~\citep{zhang2025dreamvla, lin2025hif, liu2026last, chen2026dial, luo2026being, ma2026internvla}, as illustrated in Fig.~\ref{fig:pic_method}(b).
Positioned at the end of the sequence, these queries attend across preceding visual and textual tokens to aggregate task-relevant multimodal context into compact predictive representations.
Whereas Implicit Future Alignment retains the standard cross-attention context for the action expert, this strategy explicitly augments the context with the future query tokens $x_{1:Q}^Q$ at each layer:
\begin{equation}
  \widetilde{x}_{1:K}^A = \operatorname{CrossAttn}\!\left(
    x_{1:K}^A,
    [x_{1:N}^V; x_{1:M}^L; x_{1:Q}^Q]
  \right).
  \label{eq:query-conditioned-action}
\end{equation}
To enforce world modeling on these future query tokens, their final-layer representations are fed into a lightweight decoder $D$, which reconstructs dense features to align with the target world representation $z_{t:t+T}$:
\begin{equation}
  \mathcal{L}_{\text{world}} = 1 - \cos\!\left( D(x_{1:Q}^Q), z_{t:t+T} \right).
  \label{eq:future-query-states}
\end{equation}
This predictive objective encourages the future query tokens to capture compact, task-conditioned world dynamics during training.
At inference, the dense decoder $D$ is discarded, allowing the action expert to directly leverage these learned dynamics without incurring extra predictive overhead.

\subsection{Strategy 3: MoT Future Expert}
The third strategy instantiates world modeling via a dedicated Future Expert that operates as a parallel Mixture-of-Transformers (MoT) stream alongside the VLM backbone and the Action Expert~\citep{lv2025f1, zhang2025unijepa, bi2025motus, won2025dual, ma2026internvla, hu2026bagelvla}, as illustrated in Fig.~\ref{fig:pic_method}(c).
This paradigm decouples the world modeling task from the VLM backbone, enabling the Action Expert to progressively interact with the Future Expert across all layers to leverage intermediate world dynamics.

Specifically, the Future Expert is trained from scratch as a dedicated stream operating on a set of generation tokens $x_{1:P}^G$.
These tokens are initialized with visual features extracted from the current observation $o_t$ via the frozen encoder $E$, and are progressively refined to decode the target world representation.
At each layer, the Future Expert refines these generation tokens $x_{1:P}^G$ via cross-attention exclusively with the VLM tokens, remaining isolated from the action tokens:
\begin{equation}
  \widetilde{x}_{1:P}^G = \operatorname{CrossAttn}\!\left( x_{1:P}^G, [x_{1:N}^V; x_{1:M}^L] \right).
  \label{eq:mot-future-expert}
\end{equation}
Rather than conditioning solely on the final predicted representation, the Action Expert interacts with both the VLM and the Future Expert across all intermediate layers to absorb evolving world dynamics:
\begin{equation}
  \widetilde{x}_{1:K}^A = \operatorname{CrossAttn}\!\left(
    x_{1:K}^A,
    [x_{1:N}^V; x_{1:M}^L; \widetilde{x}_{1:P}^G]
  \right).
  \label{eq:mot-action-fusion}
\end{equation}
For world model supervision, a lightweight MLP projects the final-layer generation tokens $x_{1:P}^G$ to align with the target world representation $z_{t:t+T}$:
\begin{equation}
  \mathcal{L}_{\text{world}} = 1 - \cos\!\left( \operatorname{MLP}(x_{1:P}^G), z_{t:t+T} \right).
  \label{eq:mot-world-loss}
\end{equation}
At inference, this auxiliary projection head is discarded, while the Future Expert is retained as an internal world model stream to guide action generation.

\subsection{Strategy 4: External World Predictor}
The fourth strategy, External World Predictor, decouples world modeling into a standalone module and explicitly conditions action generation on the final predicted world representation~\citep{chen2026lawam}, as illustrated in Fig.~\ref{fig:pic_method}(d).

Specifically, this strategy employs a dedicated multi-layer Transformer as the external world predictor.
The predictor operates on a sequence of generation tokens $x_{1:P}^G$ initialized with visual features from the current observation $o_t$ via the frozen encoder $E$.
To capture task-conditioned dynamics, $Q$ future query tokens $x_{1:Q}^Q$ are inserted into the VLM backbone, whose final-layer representations modulate the world predictor via Adaptive Layer Normalization (AdaLN):
\begin{equation}
  \hat{z}_{t:t+T} = \operatorname{WorldPredictor}\!\left( x_{1:P}^G, x_{1:Q}^Q \right).
  \label{eq:world-prediction}
\end{equation}
To guide action generation with the predicted world representation $\hat{z}_{t:t+T}$, the Action Expert integrates it directly into the cross-attention context:
\begin{equation}
  \widetilde{x}_{1:K}^A = \operatorname{CrossAttn}\!\left( x_{1:K}^A, \left[ x_{1:N}^V; x_{1:M}^L; \hat{z}_{t:t+T} \right] \right).
  \label{eq:world-conditioned-action}
\end{equation}
The world model supervision is applied to the external world predictor by aligning with the target world representation $z_{t:t+T}$:
\begin{equation}
  \mathcal{L}_{\text{world}} = 1 - \cos\!\left( \hat{z}_{t:t+T}, z_{t:t+T} \right).
  \label{eq:strategy4-loss}
\end{equation}
At inference, the external world predictor is retained to explicitly supply the predicted world representation $\hat{z}_{t:t+T}$ for policy execution.

\section{Experiments}

\subsection{Experimental Setup}
\label{subsec:experimental_setup}

We evaluate on LIBERO~\citep{liu2023libero} for multi-task manipulation, LIBERO-Plus~\citep{fei2025libero} for OOD robustness, RoboTwin 2.0~\citep{chen2025robotwin} for bimanual manipulation across task horizons, and DOMINO~\citep{fang2026towards} for dynamic manipulation.
The four paradigms are abbreviated as \textbf{WM-Align}, \textbf{WM-Query}, \textbf{WM-MoT}, and \textbf{WM-Ext}.
Our primary comparisons share the Qwen3-VL-2B backbone and training protocol; $\pi_{0.5}$ experiments additionally examine transfer to a pretrained VLA.
Across all paradigms, frozen encoders extract world supervision targets, and additionally provide current visual features to initialize generative tokens in WM-MoT and WM-Ext, following~\citep{zhang2025unijepa, lv2025f1, chen2026lawam}.
Implementation details and disaggregated results appear in Appendix~\ref{app:experimental_setup}.

\subsection{World Representation Analysis across Paradigms}
\label{subsec:world_representation_analysis}

\begin{figure*}[h]
    \centering
    \includegraphics[width=0.85\textwidth]{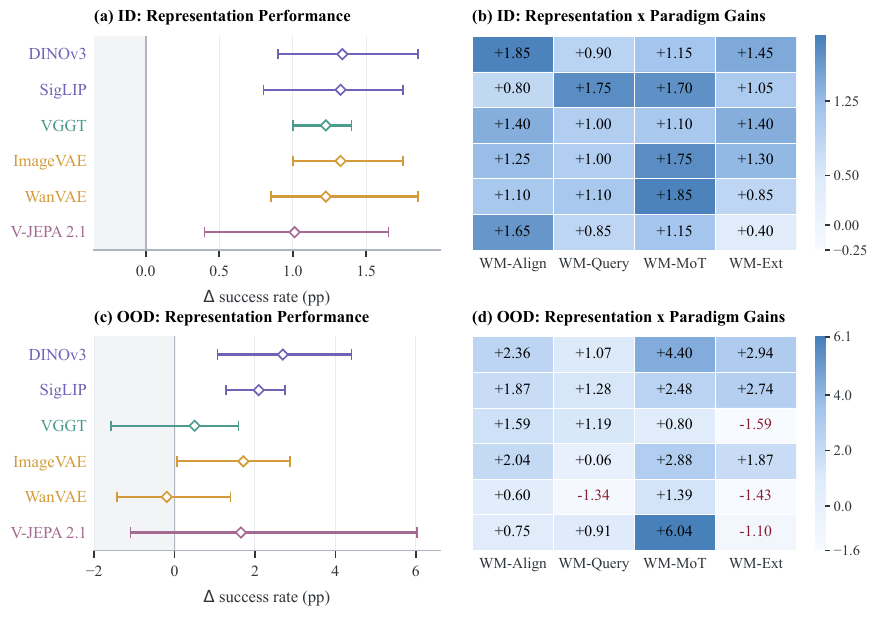}
    \caption{\textbf{World-representation performance across integration paradigms.}
    Success rate gains over the baseline on LIBERO (ID, top) and LIBERO-Plus (OOD, bottom).
    In (a, c), horizontal bars and diamonds indicate the min--max range and mean across the four paradigms, respectively.
    In (b, d), heatmaps display individual representation--paradigm gains.}
    \label{fig:extractor_stability}
    \vspace{-10pt}
\end{figure*}

\textbf{Finding 1: World representation rankings reverse across integration paradigms.}
As shown in Figure~\ref{fig:extractor_stability}, no single world representation universally dominates across architectures; instead, their relative rankings reverse across paradigms.
For example, V-JEPA~2.1 provides the strongest OOD gain for WM-MoT, yet falls below the baseline when paired with WM-Ext, where DINOv3 and SigLIP perform best.
Similarly, VGGT benefits WM-Align but degrades WM-Ext.
Furthermore, WM-Align is relatively insensitive to the representation choice, whereas WM-MoT and WM-Ext exhibit high variance, making encoder selection consequential for decoupled designs.

\textbf{Finding 2: 2D semantic representations offer robust, cross-paradigm gains.}
DINOv3 and SigLIP consistently improve both ID and OOD averages across all four paradigms.
In contrast, pixel-level representations are generally vulnerable to visual distribution shifts, causing WanVAE to underperform OOD; however, ImageVAE retains comprehensive spatial details with minimal information loss, still providing modest gains across paradigms.
Meanwhile, VGGT captures low-level spatial geometry rather than high-level task semantics, rendering its utility fragile and architecture-dependent.
Between the two semantic representations, DINOv3 and SigLIP perform comparably in-domain, whereas DINOv3 attains both the higher average gain and the higher upper bound across paradigms out-of-domain (Figure~\ref{fig:extractor_stability}(a, c)).
We therefore adopt DINOv3 as the default world representation for subsequent comparisons.

\subsection{Main Results across Manipulation Benchmarks}
\label{subsec:main_manipulation_results}

\begin{table}[h]
    \centering
    \caption{LIBERO and LIBERO-Plus performance across world modeling paradigms.
    Scores report success rates (\%), with parenthesized arrows indicating gains over the action-only baseline.
    \textbf{Bold} and \underline{underline} denote best and second-best results, respectively.}

    \label{tab:vla-robustness}
    \small
    \setlength{\tabcolsep}{3.5pt}
    \renewcommand{\arraystretch}{1.03}
    \newcommand{\upscore}[2]{%
        #1\,{\scriptsize\textcolor{green!55!black}{($\uparrow$\,#2)}}}
    \newcommand{\bestupscore}[2]{%
        \textbf{#1}\,{\scriptsize\textcolor{green!55!black}{($\uparrow$\,#2)}}}
    \newcommand{\downscore}[2]{%
        #1\,{\scriptsize\textcolor{red!70!black}{($\downarrow$\,#2)}}}
    \resizebox{0.98\linewidth}{!}{%
    \begin{tabular}{@{}lccccccccc@{}}
        \toprule
        \multirow{2}{*}{\textbf{Method}}
        & \multirow{2}{*}{\textbf{LIBERO}}
        & \multicolumn{8}{c}{\textbf{LIBERO-Plus}} \\
        \cmidrule(l){3-10}
        & & Camera & Robot & Language & Light & Background & Noise & Layout & Avg. \\
        \midrule
        OpenVLA-OFT
        & 97.1 & 56.4 & 31.9 & 79.5 & 88.7 & 93.3 & \textbf{75.8} & 74.2 & 69.6 \\
        Fast-WAM
        & 97.6 & 18.8 & 45.7 & 70.1 & 83.2 & 45.7 & 29.8 & 62.7 & 49.1 \\
        4D-WAM
        & \textbf{98.6} & 45.2 & \textbf{64.3} & 90.6 & 94.3 & 57.7 & 69.1 & \textbf{79.2} & 71.0 \\
        \midrule
        Baseline
        & 96.7 & 48.2 & 51.4 & \underline{91.4} & 96.0 & 92.9 & 65.3 & 71.3 & 71.8 \\
        \addlinespace[2pt]
        WM-Align
        & \bestupscore{98.6}{1.9}
        & \upscore{\underline{58.0}}{9.8}
        & \downscore{49.7}{1.7}
        & \downscore{89.5}{1.9}
        & \downscore{94.6}{1.4}
        & \upscore{93.5}{0.6}
        & \upscore{71.8}{6.5}
        & \upscore{73.9}{2.6}
        & \upscore{74.1}{2.3} \\
        WM-Query
        & \upscore{97.6}{0.9}
        & \upscore{54.4}{6.2}
        & \downscore{50.6}{0.8}
        & \downscore{88.5}{2.9}
        & \downscore{95.1}{0.9}
        & \bestupscore{96.3}{3.4}
        & \upscore{69.5}{4.2}
        & \downscore{69.5}{1.8}
        & \upscore{72.9}{1.1} \\
        WM-MoT
        & \upscore{97.9}{1.2}
        & \bestupscore{59.7}{11.5}
        & \upscore{\underline{53.5}}{2.1}
        & \bestupscore{91.6}{0.2}
        & \bestupscore{97.6}{1.6}
        & \upscore{\underline{95.9}}{3.0}
        & \upscore{72.3}{7.0}
        & \upscore{75.2}{3.9}
        & \bestupscore{76.2}{4.4} \\
        WM-Ext
        & \upscore{\underline{98.2}}{1.5}
        & \upscore{55.7}{7.5}
        & \downscore{46.8}{4.6}
        & \downscore{91.2}{0.2}
        & \upscore{\underline{96.8}}{0.8}
        & \upscore{95.5}{2.6}
        & \upscore{\underline{74.2}}{8.9}
        & \upscore{\underline{75.9}}{4.6}
        & \upscore{\underline{74.7}}{2.9} \\
        \bottomrule
    \end{tabular}%
    }

\end{table}

\textbf{Finding 3: World modeling improves overall performance and visual robustness.}
As shown in Table~\ref{tab:vla-robustness}, all four paradigms consistently improve success over the action-only baseline across in-domain LIBERO and out-of-domain LIBERO-Plus.
Notably, while our action-only baseline already matches or exceeds competitive references like OpenVLA-OFT~\citep{kim2025fine} and 4D-WAM~\citep{yang20264d} out-of-domain, world modeling provides substantial further gains across all paradigms.
In-domain, WM-Align reaches the highest score of 98.6\% by enriching individual visual tokens with fine-grained local dynamics through patch-wise alignment.
Across out-of-domain perturbations, performance gains are shared on camera shifts, sensor noise, and background variations, with WM-MoT providing the largest overall boost of 4.4\% and improving across all seven categories.
Controlled ablations confirm that future predictive supervision yields a larger gain than adding MoT capacity and generative token inputs alone, as detailed in Appendix~\ref{app:mot_supervision_control}.
Meanwhile, under linguistic perturbations, the decoupled WM-MoT and WM-Ext avoid notable drops, whereas coupled paradigms degrade by up to 2.9\%, suggesting that separating future prediction may mitigate interference with the VLM's pretrained language knowledge.

\begin{table}[h]
    \vspace{-4pt}
    \centering
    \caption{RoboTwin 2.0 performance under Clean2Clean and Clean2Random evaluation.
    Scores report success rates (\%), with parenthesized arrows indicating gains over the action-only baseline.
    Horizons are categorized by average demonstration length: short ($\le 200$), medium ($200$--$300$), and long ($> 300$ steps).
    \textbf{Bold} and \underline{underline} denote best and second-best results, respectively.}
    \label{tab:robotwin-c2c-c2r}
    \small
    \setlength{\tabcolsep}{4.5pt}
    \renewcommand{\arraystretch}{1.03}
    \newcommand{\upscore}[2]{%
        #1\,{\scriptsize\textcolor{green!55!black}{($\uparrow$\,#2)}}}
    \newcommand{\bestupscore}[2]{%
        \textbf{#1}\,{\scriptsize\textcolor{green!55!black}{($\uparrow$\,#2)}}}
    \newcommand{\downscore}[2]{%
        #1\,{\scriptsize\textcolor{red!70!black}{($\downarrow$\,#2)}}}
    \resizebox{0.98\linewidth}{!}{%
    \begin{tabular}{@{}lcccccccc@{}}
        \toprule
        \multirow{2}{*}{\textbf{Method}}
        & \multicolumn{4}{c}{\textbf{Clean2Clean}}
        & \multicolumn{4}{c}{\textbf{Clean2Random}} \\
        \cmidrule(lr){2-5}\cmidrule(l){6-9}
        & Short & Medium & Long & Avg. & Short & Medium & Long & Avg. \\
        \midrule
        starVLA
        & 49.7 & 51.9 & 28.8 & 46.5
        & 4.0 & 3.2 & 0.6 & 3.2 \\
        AHA-WAM
        & 60.3 & \underline{76.7} & \textbf{59.1} & \underline{64.3}
        & 2.6 & 5.8 & 1.2 & 3.2 \\
        GalaxeaVLA
        & \textbf{63.0} & 71.6 & 48.9 & 62.7
        & 11.5 & \underline{19.6} & 6.7 & 12.7 \\
        \midrule
        Baseline & 53.4 & 65.0 & 30.7 & 52.3 & \underline{13.4} & 15.4 & 6.6 & 12.7 \\
        \addlinespace[2pt]
        WM-Align
        & \upscore{\underline{62.0}}{8.6}
        & \upscore{72.4}{7.4}
        & \upscore{48.8}{18.1}
        & \upscore{62.3}{10.0}
        & \bestupscore{15.3}{1.9}
        & \upscore{17.9}{2.5}
        & \upscore{8.9}{2.3}
        & \bestupscore{14.8}{2.1} \\
        WM-Query
        & \upscore{61.1}{7.7}
        & \bestupscore{77.8}{12.8}
        & \upscore{\underline{57.0}}{26.3}
        & \bestupscore{64.7}{12.4}
        & \downscore{11.9}{1.5}
        & \upscore{15.8}{0.4}
        & \bestupscore{10.3}{3.7}
        & \downscore{12.6}{0.1} \\
        WM-MoT
        & \upscore{60.3}{6.9}
        & \upscore{74.1}{9.1}
        & \upscore{52.9}{22.2}
        & \upscore{62.6}{10.3}
        & \downscore{12.6}{0.8}
        & \bestupscore{20.1}{4.7}
        & \upscore{\underline{9.8}}{3.2}
        & \upscore{\underline{14.0}}{1.3} \\
        WM-Ext
        & \upscore{58.9}{5.5}
        & \upscore{73.8}{8.8}
        & \upscore{47.2}{16.5}
        & \upscore{60.7}{8.4}
        & \downscore{10.9}{2.5}
        & \upscore{16.4}{1.0}
        & \upscore{7.9}{1.3}
        & \downscore{11.8}{0.9} \\
        \bottomrule
    \end{tabular}%
    }
    \vspace{-4pt}

\end{table}

\textbf{Finding 4: World modeling yields substantial gains on long-horizon tasks.}
As reported in Table~\ref{tab:robotwin-c2c-c2r}, world modeling consistently enhances RoboTwin 2.0 manipulation performance, especially as task horizons extend.
Under Clean2Clean, all paradigms achieve notable improvements that progressively increase with horizon length, yielding gains of 5.5\% to 8.6\% on short tasks, 7.4\% to 12.8\% on medium tasks, and 16.5\% to 26.3\% on long tasks.
Under Clean2Random, all four paradigms improve long-horizon success by 1.3\% to 3.7\%. In contrast, on short tasks, only WM-Align improves by 1.9\%, while the other three drop by 0.8\% to 2.5\%. This indicates that short tasks rely primarily on immediate visual feedback, where predicting future states offers little benefit and can even introduce slight noise under visual shifts.

\subsection{Performance under Dynamic Environments and Data Scarcity}
\label{subsec:performance_dynamic_data_scarcity}

\begin{wrapstuff}[r,width=0.46\textwidth,type=table,abovesep=-0.6\baselineskip]
    \centering
    \captionof{table}{Performance on DOMINO Level~1 dynamic manipulation.
    Scores report success rate (SR, \%) and manipulation score (MS).}
    \label{tab:dynamic-environments}
    \small
    \setlength{\tabcolsep}{2pt}
    \renewcommand{\arraystretch}{1.03}
    \newcommand{\upscore}[2]{%
        #1\,{\scriptsize\textcolor{green!55!black}{($\uparrow$\,#2)}}}
    \newcommand{\bestupscore}[2]{%
        \textbf{#1}\,{\scriptsize\textcolor{green!55!black}{($\uparrow$\,#2)}}}
    \begin{tabular*}{\linewidth}{@{\extracolsep{\fill}}lcc@{}}
        \toprule
        \textbf{Method} & \textbf{SR} & \textbf{MS} \\
        \midrule
        InternVLA-M1  & 5.4 & 27.6 \\
        Isaac-GR00T   & 6.1 & 28.6 \\
        OpenVLA-OFT   & 9.1 & 24.1 \\
        $\pi_{0.5}$   & 9.6 & 26.2 \\
        \midrule
        Baseline & 10.6 & 29.1 \\
        \addlinespace[2pt]
        WM-Align
        & \upscore{11.9}{1.3}
        & \upscore{30.1}{1.0} \\
        WM-Query
        & \upscore{13.5}{2.9}
        & \upscore{32.0}{2.9} \\
        WM-MoT
        & \bestupscore{15.5}{4.9}
        & \upscore{\underline{32.9}}{3.8} \\
        WM-Ext
        & \upscore{\underline{15.1}}{4.5}
        & \bestupscore{33.0}{3.9} \\
        \bottomrule
    \end{tabular*}
\end{wrapstuff}

\noindent
\textbf{Finding 5: Decoupled predictive designs achieve stronger gains in dynamic manipulation.}
Table~\ref{tab:dynamic-environments} reports DOMINO Level~1 results alongside representative VLAs, including InternVLA-M1~\citep{chen2025internvla}, Isaac-GR00T~\citep{bjorck2025gr00t}, OpenVLA-OFT~\citep{kim2025fine}, and $\pi_{0.5}$~\citep{intelligence2025pi}.
While all four paradigms outperform prior models and the 10.6\% action-only baseline, their gains vary across architectures.
WM-Align and WM-Query achieve 11.9\% and 13.5\% success, respectively, whereas decoupled predictive designs such as WM-Ext and WM-MoT reach 15.1\% and 15.5\% success.
One possible explanation is that enforcing token-wise patch alignment or bottleneck queries on shared representations provides limited capacity for dynamic modeling, whereas dedicated predictive modules can more flexibly capture continuous object motion.
\par\medskip

\begin{figure*}[htbp]
    \vspace{-8pt}
    \centering
        \includegraphics[width=0.72\linewidth]{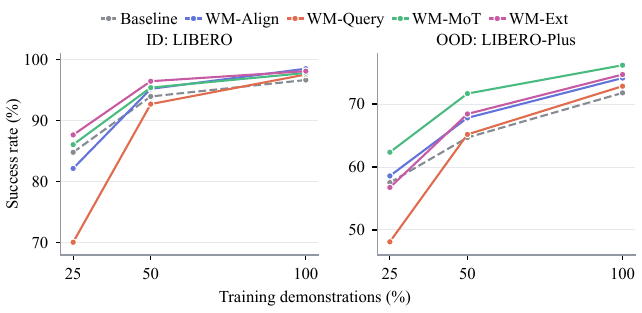}
        \vspace{-3mm}
        \caption{\textbf{Data scaling on LIBERO and LIBERO-Plus.}
        Absolute success rates using 25\%, 50\%, and 100\% of the training
        demonstrations.}
        \label{fig:data_scaling}
        \vspace{-8pt}
    \end{figure*}

\textbf{Finding 6: Decoupled paradigms retain stronger performance under data scarcity.}
We compare policies trained with 25\%, 50\%, and 100\% demonstrations on LIBERO and LIBERO-Plus, as illustrated in Figure~\ref{fig:data_scaling} and detailed in Tables~\ref{tab:data-scaling-id-details}--\ref{tab:data-scaling-ood-details}.
Overall, decoupled architectures such as WM-Ext and WM-MoT maintain strong performance under data scarcity.
In contrast, WM-Query degrades sharply in low-data settings, likely because the added query tokens require sufficient demonstrations to learn compact and effective world representations.
Between the two decoupled designs, their relative advantages differ across domains.
Specifically, WM-Ext achieves the highest in-domain sample efficiency, reaching 96.5\% success at 50\% data, while WM-MoT provides the strongest out-of-distribution robustness, leading all alternatives with 62.4\% success at 25\% data.

\subsection{Compatibility with Pre-trained Foundation VLAs}
\label{subsec:compatibility_pretrained_vla}

\textbf{Finding 7: Decoupled designs show better compatibility with the pretrained VLA.}
To examine how world modeling integrates into foundation VLAs, we incorporate all four paradigms into $\pi_{0.5}$ and evaluate them on RoboTwin 2.0 under Clean2Clean and Clean2Random settings, as reported in Table~\ref{tab:pi05-robotwin-c2c-c2r}.
Decoupled architectures transfer more favorably to the pretrained backbone, as both WM-MoT and WM-Ext improve overall success rates under in-domain and out-of-distribution conditions.
Between the two, WM-Ext achieves the largest in-domain gain, while WM-MoT provides the strongest out-of-distribution robustness.
In contrast, coupled designs do not exhibit consistent gains.
Specifically, WM-Query performs slightly below the baseline, while WM-Align degrades policy performance, dropping success rates by over 5\% across both settings.
A plausible reason is that WM-Align imposes auxiliary predictive objectives directly onto the shared visual representations, which can distort the pretrained feature space and disrupt the manipulation priors established during pretraining.

\begin{table}[h]
    \vspace{-8pt}

    \centering
    \caption{$\pi_{0.5}$ performance on RoboTwin 2.0 under Clean2Clean and Clean2Random evaluation.
    Scores report success rates (\%), with parenthesized arrows indicating gains over the $\pi_{0.5}$ baseline.
    \textbf{Bold} and \underline{underline} denote best and second-best results, respectively.}
    \label{tab:pi05-robotwin-c2c-c2r}
    \small
    \setlength{\tabcolsep}{4.5pt}
    \renewcommand{\arraystretch}{1.03}
    \newcommand{\piupscore}[2]{%
        #1\,{\scriptsize\textcolor{green!55!black}{($\uparrow$\,#2)}}}
    \newcommand{\pidownscore}[2]{%
        #1\,{\scriptsize\textcolor{red!70!black}{($\downarrow$\,#2)}}}
    \resizebox{0.98\linewidth}{!}{%
    \begin{tabular}{@{}lcccccccc@{}}
        \toprule
        \multirow{2}{*}{\textbf{Method}}
        & \multicolumn{4}{c}{\textbf{Clean2Clean}}
        & \multicolumn{4}{c}{\textbf{Clean2Random}} \\
        \cmidrule(lr){2-5}\cmidrule(l){6-9}
        & Short & Medium & Long & Avg. & Short & Medium & Long & Avg. \\
        \midrule
        $\pi_{0.5}$ & 67.9 & 72.5
        & \underline{68.2} & 69.1 & 45.6
        & \textbf{53.3} & 39.0 & 46.4 \\
        \addlinespace[2pt]
        $\pi_{0.5}$ + WM-Align
        & \pidownscore{64.1}{3.8}
        & \pidownscore{66.8}{5.7}
        & \pidownscore{57.6}{10.6}
        & \pidownscore{63.6}{5.5}
        & \pidownscore{39.9}{5.7}
        & \pidownscore{47.2}{6.1}
        & \pidownscore{35.9}{3.1}
        & \pidownscore{41.0}{5.4} \\
        $\pi_{0.5}$ + WM-Query
        & \pidownscore{67.8}{0.1}
        & \pidownscore{72.4}{0.1}
        & \pidownscore{66.0}{2.2}
        & \pidownscore{68.7}{0.4}
        & \pidownscore{45.4}{0.2}
        & \pidownscore{50.7}{2.6}
        & \piupscore{42.3}{3.3}
        & \pidownscore{46.2}{0.2} \\
        $\pi_{0.5}$ + WM-MoT
        & \piupscore{\underline{69.6}}{1.7}
        & \piupscore{\textbf{76.8}}{4.3}
        & \piupscore{\textbf{68.6}}{0.4}
        & \piupscore{\underline{71.3}}{2.2}
        & \piupscore{\textbf{49.6}}{4.0}
        & \pidownscore{51.5}{1.8}
        & \piupscore{\textbf{47.2}}{8.2}
        & \piupscore{\textbf{49.7}}{3.3} \\
        $\pi_{0.5}$ + WM-Ext
        & \piupscore{\textbf{71.5}}{3.6}
        & \piupscore{\underline{74.9}}{2.4}
        & \pidownscore{67.9}{0.3}
        & \piupscore{\textbf{71.7}}{2.6}
        & \piupscore{\underline{48.1}}{2.5}
        & \pidownscore{\underline{52.4}}{0.9}
        & \piupscore{\underline{44.1}}{5.1}
        & \piupscore{\underline{48.5}}{2.1} \\
        \bottomrule
    \end{tabular}%
    }
    \vspace{-8pt}
\end{table}

\subsection{Real World Experiments}
\label{subsec:real_world_experiments}

\begin{figure*}[htbp]
    \vspace{-8pt}
    \centering
    \includegraphics[width=0.95\textwidth]{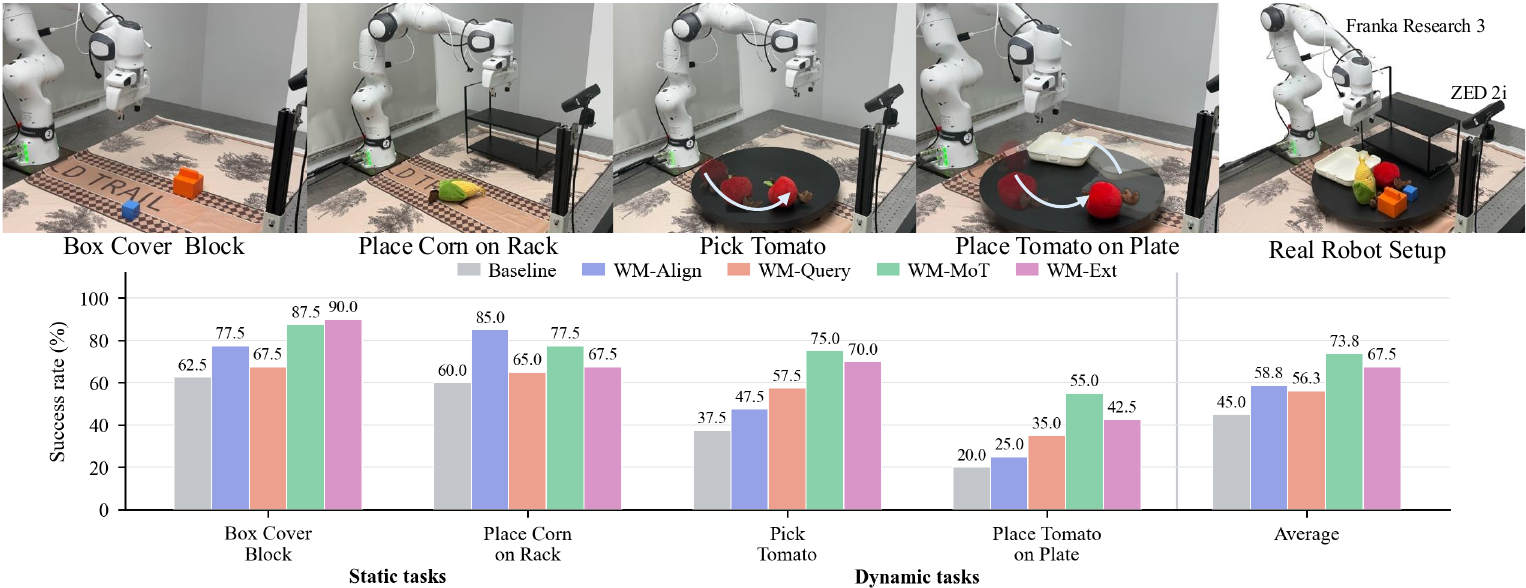}
    \caption{\textbf{Real-world experimental setup and manipulation tasks.}}
    \label{fig:realworld}
    \vspace{-8pt}
\end{figure*}

\textbf{Finding 8: World modeling consistently improves real-world manipulation performance.}
Across the four real-world tasks illustrated in Figure~\ref{fig:realworld}, all four world modeling paradigms consistently outperform the action-only baseline, raising the average success rate from 45.0\% to 56.3--73.8\%, with detailed experimental setups provided in Appendix~\ref{app:experimental_setup} and full per-task numerical results in Table~\ref{tab:app-real-robot-per-task} (Appendix~\ref{app:real_robot_details}).
On static tasks, all paradigms achieve steady improvements over the baseline, with WM-Ext achieving 90.0\% on box cover block and WM-Align reaching 85.0\% on placing corn.
On dynamic tasks, performance gains differ noticeably across architectures. While WM-Align yields modest improvements, decoupled designs such as WM-MoT and WM-Ext provide substantial gains. Specifically, WM-MoT reaches 75.0\% on dynamic picking and 55.0\% on tomato placement, achieving the highest overall average of 73.8\%.

\section{Conclusion}

In this paper, we presented a systematic empirical investigation of world modeling in Vision-Language-Action policies across diverse world representations, four representative integration paradigms, and distinct policy backbones.
First, 2D semantic representations provide reliable gains across paradigms, whereas pixel and 3D representations are more sensitive to integration architectures.
Second, integration paradigms present distinct trade-offs, where coupled designs offer compact representations for static manipulation, while decoupled architectures provide superior robustness under distribution shifts and dynamic manipulation.
Finally, decoupled integration shows better compatibility when adapting pretrained foundation VLAs.
We hope our unified framework and design principles provide valuable guidance for future research on integrating world modeling into robot foundation models.

\section*{LLM Usage Disclosure}
In this work, generative AI and LLMs were used solely to aid and polish paper writing. The authors have reviewed and edited all content and take full responsibility for the final submission.

\bibliography{iclr2027_conference}
\bibliographystyle{iclr2027_conference}

\appendix
\counterwithin{table}{section}
\renewcommand{\theHtable}{appendix.\thesection.\arabic{table}}
\clearpage
\section*{Appendix Overview}

The appendix is organized into three parts:
\begin{description}[leftmargin=2.2em,labelsep=0.6em,style=multiline,font=\normalfont\bfseries]
  \item[\ref{app:experimental_setup}]
  \textbf{Experimental Setup and Implementation Details.}
  Benchmarks, representation encoders, policy backbones, training specifications,
  evaluation protocols, and computational costs.

  \item[\ref{app:controls}]
  \textbf{Controlled Ablations, Robustness, and Diagnostic Analyses.}
  Multi-seed replications, controlled supervision ablations, sensitivity sweeps, and
  methodological scope of claims.

  \item[\ref{app:tables}]
  \textbf{Detailed Benchmark Results and Per-Task Breakdowns.}
  Complete results for LIBERO, LIBERO-Plus, RoboTwin, DOMINO, and real-world robot manipulation.
\end{description}

\section{Experimental Setup and Implementation Details}
\label{app:experimental_setup}

This section provides comprehensive specifications for the simulation benchmarks, target world representation encoders, policy backbones and world modeling integration paradigms, training hyperparameter configurations, evaluation protocols, and computational and deployment costs.

\subsection{Benchmarks and Evaluation Tasks}
\label{app:benchmarks}

We evaluate all world modeling strategies and baselines across four simulation benchmarks (LIBERO, LIBERO-Plus, RoboTwin 2.0, and DOMINO) and a physical Franka Research 3 robotic setup:

\paragraph{LIBERO.}
LIBERO~\citep{liu2023libero} is a widely adopted benchmark designed for multi-task robotic manipulation, featuring procedural tabletop scenes in MuJoCo. We evaluate policies across all four official benchmark suites:
\begin{itemize}[leftmargin=*,noitemsep,topsep=2pt]
  \item \textit{Spatial} (10 tasks): Evaluates transferring knowledge of spatial relationships, where tasks share the same set of objects but require distinguishing identical objects based on their locations and spatial configurations.
  \item \textit{Object} (10 tasks): Evaluates transferring knowledge of object categories, where tasks require manipulating distinct object types with varied geometries and visual appearances.
  \item \textit{Goal} (10 tasks): Evaluates transferring knowledge of task goals, where tasks share identical objects and spatial layouts but require executing different motions and behavioral goals specified by language.
  \item \textit{Long} (10 tasks): Evaluates long-horizon manipulation requiring sequential decision-making and multi-stage skill execution.
\end{itemize}

\paragraph{LIBERO-Plus.}
To evaluate out-of-distribution (OOD) robustness under realistic distribution shifts, we benchmark on \textbf{LIBERO-Plus}~\citep{fei2025libero}, which introduces seven perturbation dimensions to the LIBERO environments:
(1)~\textit{Camera Viewpoints}: shifts in camera distance, spherical position, and orientation;
(2)~\textit{Robot Initial States}: perturbations to the arm's initial joint angles ($q_{\text{pos}}$);
(3)~\textit{Language Instructions}: LLM-based rewrites with contextual distractions and paraphrasing;
(4)~\textit{Light Conditions}: variations in lighting directions, diffuse colors, and specular highlights;
(5)~\textit{Background Textures}: randomized surface and environment textures;
(6)~\textit{Sensor Noise}: optical and photometric distortions such as Gaussian blur and fog; and
(7)~\textit{Object Layout}: unseen distractor objects and target pose variations.

\paragraph{RoboTwin 2.0.}
RoboTwin 2.0~\citep{chen2025robotwin} is a dual-arm tabletop manipulation platform.
All policies are trained on 50 clean expert demonstrations per task. Following the official benchmark protocol, evaluation is conducted with 100 rollouts per task under two evaluation settings: (1)~\textit{Clean2Clean} (C2C, corresponding to the Easy setting), where policies trained on clean data are evaluated in clean environments; and (2)~\textit{Clean2Random} (C2R, corresponding to the Hard setting), where policies trained on clean data are evaluated in domain-randomized environments (featuring variations in clutter, lighting, textures, and tabletop height) to test out-of-distribution robustness.
To diagnose compounding errors across varying temporal scales, tasks are categorized into three horizon tiers based on average demonstration completion steps: \textit{Short Horizon} ($\le 200$ steps), \textit{Medium Horizon} ($200$--$300$ steps), and \textit{Long Horizon} ($> 300$ steps).

\paragraph{DOMINO.}
DOMINO~\citep{fang2026towards} benchmarks dynamic manipulation in non-stationary environments.
We evaluate on Level~1 under the clean dynamic regime, where target objects continuously move across the table, requiring policies to anticipate future target positions for successful interaction.

\paragraph{Real-World Setup.}
Our physical experiments are conducted on a single-arm Franka Research 3 equipped with a third-person ZED 2i stereo camera.
Visual observations are supplied by the camera, and actions are executed as end-effector deltas along with an absolute gripper command.
We evaluate four manipulation tasks covering both static and dynamic settings, including two static tasks (\textit{Box Cover Block} and \textit{Place Corn on Rack}) and two dynamic tasks (\textit{Pick Tomato} and \textit{Place Tomato on Plate}).
For each task, we collect 50 demonstrations for training.
During evaluation, we run 40 trials per setting and report the average success rate.

\subsection{Target World Representation Encoders}
\label{app:encoders}

To supervise predictive dynamics across all paradigms, all six encoders construct targets using the same four future timesteps ($t+4$, $t+8$, $t+12$, and $t+16$). Image encoders process these frames independently, whereas video encoders encode them jointly over the identical prediction horizon.

We investigate six representative frozen target encoders $E$ spanning four distinct representation classes:
\begin{itemize}[leftmargin=*,noitemsep,topsep=2pt]
  \item \textbf{ImageVAE}~\citep{tian2024visual} \textit{(Pixel-level)}: A multi-scale visual autoregressive VAE trained for generative image modeling, producing dense spatial latent grids.
  \item \textbf{WanVAE}~\citep{wan2025wan} \textit{(Pixel-level)}: The 3D video autoencoder from Wan2.1, compressing spatio-temporal video chunks to capture continuous motion and dense visual appearance.
  \item \textbf{DINOv3}~\citep{simeoni2025dinov3} \textit{(2D Semantic)}: A self-supervised vision transformer capturing patch-level semantic correspondence and object boundaries, using ViT-B/16 pooled to an $8\times 8$ grid.
  \item \textbf{SigLIP}~\citep{zhai2023sigmoid} \textit{(2D Semantic)}: A language-image pre-trained transformer encoding high-level visual semantics aligned with language instructions.
  \item \textbf{VGGT}~\citep{wang2025vggt} \textit{(3D Geometric)}: A feedforward geometry transformer predicting 3D point maps and camera poses, yielding spatially structured geometric tokens.
  \item \textbf{V-JEPA~2.1}~\citep{mur2026vjepa21} \textit{(Spatiotemporal)}: A joint-embedding predictive architecture trained on video sequences by predicting masked latent features, providing motion-aware dynamic representations.
\end{itemize}

\paragraph{Design Rationale on Input and Target Pairing.}
In our cross-paradigm comparisons, WM-Align and WM-Query receive standard visual tokens from the VLM and interact with encoder $E$ solely through the future auxiliary loss.
In contrast, WM-MoT and WM-Ext utilize $E$ both to extract current observation features for generation token initialization and to construct future supervision targets.
This configuration follows established world modeling practices such as UniJEPA and LAWAM, where a predictive world model naturally operates within a homogeneous semantic feature space from current states to future transitions.
Consequently, swapping encoder $E$ in decoupled architectures alters both the current input conditioning space and the future predictive target.
The observed representation rankings in Section~\ref{subsec:world_representation_analysis} thus characterize the holistic system-level compatibility of each representation space when serving as both the current perceptual grounding and the future predictive target, rather than isolating future supervision alone.

\subsection{Policy Backbones and Baseline Implementation}
\label{app:backbones}

\paragraph{Primary VLA Testbed (Qwen3-VL-2B + Cross-DiT).}
Our primary base policy pairs a pretrained Qwen3-VL-2B~\citep{bai2025qwen3vl} vision-language model with a layer-wise Diffusion Transformer (cross-DiT) Action Expert~\citep{black2024pi0}:
\begin{itemize}[leftmargin=*,noitemsep,topsep=2pt]
  \item \textit{Visual and Language Encoding}: Multi-view camera inputs ($224 \times 224$) are embedded by Qwen3-VL's ViT into $N$ visual tokens, while task instructions are embedded into $M$ language tokens.
  \item \textit{Action Expert}: Continuous action chunks are parameterized by $K$ action tokens $x^A_{1:K}$. The Action Expert comprises 28 Transformer layers matching the depth of the VLM backbone. At each layer, per-layer projections map the multimodal context into the Action DiT latent space for cross-attention.
  \item \textit{Flow Matching Objective}: The policy is optimized using standard conditional flow matching with optimal transport paths:
  \begin{equation}
    \mathcal{L}_{\text{action}} = \mathbb{E}_{t, a_0, a_1} \left\| v_\theta(a_t, t, c) - (a_1 - a_0) \right\|^2,
  \end{equation}
  where $a_0 \sim \mathcal{N}(0, I)$, $a_1$ is the ground-truth action chunk, and $a_t = t a_1 + (1 - t) a_0$.
\end{itemize}

\paragraph{Pretrained Foundation VLA ($\pi_{0.5}$).}
$\pi_{0.5}$~\citep{intelligence2025pi} is a 3B parameter foundation VLA pre-trained on diverse web-scale robotic demonstrations. It integrates a Vision-Language-Action backbone with flow-matching action decoding. When fine-tuning with world modeling paradigms, we preserve its original tokenization and action projection configurations.

\paragraph{Implementation of World Modeling Strategies.}
\begin{itemize}[leftmargin=*,noitemsep,topsep=2pt]
  \item \textbf{Implicit Future Alignment (WM-Align, Strategy~1)}:
  Because Qwen3-VL and $\pi_{0.5}$ have different backbone depths, we attach
  supervision to a backbone-specific middle-to-late layer rather than forcing
  the same absolute index: layer~18 for Qwen3-VL and layer~12 for
  $\pi_{0.5}$. These locations are fixed across tasks and evaluation
  conditions. A two-layer MLP projection head maps the selected visual patch tokens to the target feature space of $E$. The cosine similarity loss $\mathcal{L}_{\text{world}}$ is computed per patch and averaged across all spatial tokens. The projection head is removed during inference.
  \item \textbf{Explicit Future Query (WM-Query, Strategy~2)}: $Q = 16$ learnable future query tokens $x_{1:Q}^Q$ are appended to the input token sequence as standalone parameters. At each layer, they attend to all preceding visual and language tokens. A cross-attention decoder $D$ reconstructs spatial future representations from the final-layer query representations to compute $\mathcal{L}_{\text{world}}$. During inference, $D$ is discarded and the Action DiT directly cross-attends to the learned future queries alongside visual and language tokens.
  \item \textbf{MoT Future Expert (WM-MoT, Strategy~3)}: The Future Expert is implemented as a 28-layer Transformer stream trained from scratch. Generation tokens $x_{1:P}^G$ ($P = 64$) are initialized from current observation features extracted by encoder $E$. At each layer, they perform self-attention and cross-attend to the VLM visual and text tokens. The Action Expert cross-attends to the concatenated context $[x^V; x^L; \widetilde{x}^G]$. A two-layer prediction MLP maps the final-layer generation tokens to the target feature space for $\mathcal{L}_{\text{world}}$.
  \item \textbf{External World Predictor (WM-Ext, Strategy~4)}: Employs a standalone 12-layer Transformer decoder conditioned via adaptive layer normalization (AdaLN). The $Q = 8$ query representations from the VLM backbone are projected via a two-layer conditioning MLP to modulate the AdaLN layers. The decoder iteratively updates the visual features of the current observation $o_t$ to predict future representation $\hat{z}_{t:t+T}$, which is concatenated into the cross-attention context of the Action Expert.
\end{itemize}

\subsection{Hyperparameter and Training Specifications}
\label{app:hyperparameters}

Table~\ref{tab:app-hyperparams} summarizes the Qwen-based LIBERO
configuration used for the main DINOv3 comparison and the cost measurements
in Table~\ref{tab:app-compute-cost}. We detail this configuration explicitly, as
RoboTwin, DOMINO, and $\pi_{0.5}$ adopt benchmark-specific action horizons and
batch sizes. All LIBERO models in this comparison are trained on eight NVIDIA H100 GPUs with identical optimization schedules.

\begin{table}[htbp]
  \centering
  \caption{Hyperparameters and architectural configurations for the Qwen-based LIBERO main comparison and compute-cost measurements.}
  \label{tab:app-hyperparams}
  \small
  \setlength{\tabcolsep}{6pt}
  \renewcommand{\arraystretch}{1.08}
  \begin{tabular}{@{}ll@{}}
    \toprule
    \textbf{Hyperparameter / Component} & \textbf{Specification} \\
    \midrule
    \multicolumn{2}{@{}l}{\textit{Training \& Optimization}} \\
    Optimizer & AdamW ($\beta_1 = 0.9$, $\beta_2 = 0.95$, $\epsilon=10^{-8}$) \\
    Learning rate schedule & Cosine annealing with 5,000-step linear warmup \\
    Peak learning rate & $1\times 10^{-5}$ (VLM), $1\times 10^{-4}$ (action/future modules) \\
    Batch size & 16 per GPU; 128 global (8 GPUs, no accumulation) \\
    Precision & Mixed BF16 \\
    Observation resolution & $224 \times 224$ pixels \\
    Action chunk size ($K$) & 8 \\
    Future prediction horizon ($T$) & 16 timesteps \\
    World loss weight ($\lambda_{\text{world}}$) & 0.1 \\
    Diffusion steps (training / inference) & Continuous flow matching / 4 Euler ODE steps \\
    Training iterations & 50,000 steps \\
    Hardware & $8 \times$ NVIDIA H100 (80GB SXM5) \\
    \midrule
    \multicolumn{2}{@{}l}{\textit{Architectural Configurations}} \\
    Action Expert & 28 layers, hidden dim 1024, 16 heads \\
    WM-Align Projection Head & 2-layer MLP, hidden dim 4096, GELU \\
    WM-Query Decoder $D$ & 8-head cross-attention, hidden dim 512, $Q=16$ queries \\
    WM-MoT Future Expert & 28 layers, hidden dim 512, 8 heads, 2-layer MLP (dim 2048) \\
    WM-Ext World Predictor & 12 layers, hidden dim 1024, 16 heads, AdaLN (2-layer MLP) \\
    \bottomrule
  \end{tabular}
\end{table}

\paragraph{Shared Training Protocol as a Controlled Design Choice.}
Our study evaluates how different representation-integration pairs perform
under a common training protocol. Within each controlled comparison, the
world-supervised variants share the cosine world-loss formulation, its
loss weight, future-frame sampling, and optimization schedule. Holding these
hyperparameters fixed ensures that each architecture's intrinsic compatibility
with the auxiliary objective is directly evaluated. Because distinct
architectures route predictive gradients through different submodules and target
representations possess varying feature geometries, these cross-paradigm
dynamics constitute a central focus of our analysis. The resulting rankings
thus characterize relative performance under a unified protocol,
precluding confounding factors from disparate tuning schedules.

\paragraph{Random Seeds and Controlled Experimental Protocol.}
To maintain controlled comparisons across architectural and representation choices, our primary benchmarks are trained under a unified random seed that fixes the demonstration sampling order, batch sequencing, and weight initialization, while sharing identical optimization schedules across all methods.
Evaluation rollouts are similarly executed with identical, deterministic environment seed sequences across all evaluated models.
Given the extensive rollout volume of our evaluation suite, including 10,030 closed-loop rollouts per model on LIBERO-Plus and 2,000 on LIBERO, each evaluated checkpoint is measured with high empirical precision.
To further verify that the observed architectural trends generalize beyond training stochasticity, Appendix~\ref{app:multi_seed_verification} complements the primary benchmarks with multi-seed training replications across central findings, confirming the consistent recurrence of key phenomena (such as the out-of-domain degradation of WM-Ext with V-JEPA).

\subsection{Evaluation Protocol and Metrics}
\label{app:evaluation_protocol}

For each benchmark, all methods are evaluated using checkpoints from the same training step, while the training step may differ across benchmarks. For each model, the same checkpoint is used for both in-domain and out-of-domain evaluation.
All simulated and real-world experiments follow standardized closed-loop rollout protocols:
\begin{itemize}[leftmargin=*,noitemsep,topsep=2pt]
  \item \textbf{LIBERO and LIBERO-Plus}: LIBERO uses 50 independent episodes per task (2,000 episodes across 40 tasks), while LIBERO-Plus uses 10,030 episodes per model across the seven perturbation axes. We record the binary Success Rate (SR, \%) per task and compute suite-level and perturbation-level averages. Perturbation categories in LIBERO-Plus contain different numbers of episodes. Following prior work, the overall average success rate is computed as $100 \times \frac{\sum_{c=1}^{7} S_c}{\sum_{c=1}^{7} N_c}$, where $S_c$ and $N_c$ denote the numbers of successful and total episodes in category $c$, respectively.
  \item \textbf{RoboTwin 2.0}: Evaluated over 100 rollouts per task under each of the Clean2Clean (C2C) and Clean2Random (C2R) conditions. Performance is stratified across Short ($\le 200$), Medium ($200$--$300$), and Long ($> 300$) completion step categories. We report task-level SR (\%) and averages across tasks within each horizon group and overall.
  \item \textbf{DOMINO}: Tested across 100 episodes per task on Level~1. We report Success Rate (SR, \%) and Manipulation Score (MS, $[0, 100]$), which provides a fine-grained continuous metric measuring intermediate task progression even in the event of terminal task failure.
  \item \textbf{Real-World Robot}: Real-world evaluations follow the physical setup described in Section~\ref{subsec:experimental_setup}, with 40 independent trials per task. We report per-task SR (\%) along with static, dynamic, and overall averages in Table~\ref{tab:app-real-robot-per-task}.
\end{itemize}

\subsection{Compute and Deployment Cost}
\label{app:compute_cost}

To accurately reflect computational overhead across paradigms rather than equating training steps to identical compute, Table~\ref{tab:app-compute-cost} reports empirical training and deployment measurements. We measure 50k-step wall-clock time on
an identical server node equipped with eight NVIDIA H100 (80GB SXM5) GPUs and a per-GPU batch size of 16.
Policy latency is benchmarked using one 224$\times$224 RGB observation, BF16
policy weights, the configured four flow-matching steps, 5 warm-up iterations,
and 30 timed runs with CUDA synchronization on a single H100 GPU. Online DINOv3 encoding is measured separately
in FP32 over 50 timed iterations following 5 warm-up steps. Training time excludes
one-time DINOv3 feature precomputation, which is shared by the four
world-supervised variants in this comparison and amortized across runs.
The main LIBERO success rates use the same action horizon, four-step Euler
solver, and DINOv3 ViT-B/16 encoder, so the accuracy and latency numbers
correspond to the same deployment configuration. All policy latencies are measured on
architecture-matched checkpoints, ensuring identical computation graphs and tensor profiles.

\textit{Training parameters} account for all modules updated during optimization, whereas \textit{deployment parameters} exclude the auxiliary decoders of WM-Align and WM-Query (which are discarded at inference) and incorporate the frozen 85.7M DINOv3 encoder required online by WM-MoT and WM-Ext. Consequently,
WM-Align returns to the baseline parameter count and latency after training.
WM-Query also has near-baseline deployment cost, while retaining its future
query tokens in the VLM.

\begin{table*}[t]
  \centering
  \caption{Computational and deployment efficiency across world modeling paradigms ($8\times$ NVIDIA H100). Training time is measured over 50k steps. Deployment parameters include the frozen DINOv3 encoder (85.7M) when required. Latencies are per-step batch-1 medians: ``Policy'' measures policy forward pass, ``Encode'' isolates online visual feature extraction, and ``E2E'' denotes end-to-end latency.}
  \label{tab:app-compute-cost}
  \scriptsize
  \setlength{\tabcolsep}{5.0pt}
  \renewcommand{\arraystretch}{1.15}
  \begin{tabular*}{\textwidth}{@{\extracolsep{\fill}}lcccccc@{}}
    \toprule
    \textbf{Method}
    & \textbf{Train Params.}
    & \textbf{50k Train Time}
    & \textbf{Deploy Params.}
    & \textbf{Policy}
    & \textbf{Encode}
    & \textbf{E2E} \\
    & \textbf{(B)} & \textbf{(h [GPU-h])} & \textbf{(B)}
    & \textbf{(ms)} & \textbf{(ms)} & \textbf{(ms)} \\
    \midrule
    Baseline & 2.608 & 8.71 (69.7) & 2.608 & 103.3 & -- & 103.3 \\
    WM-Align & 2.629 & 9.13 (73.0) & 2.608 & 103.0 & -- & 103.0 \\
    WM-Query & 2.613 & 9.04 (72.3) & 2.608 & 105.4 & -- & 105.4 \\
    WM-MoT & 2.866 & 10.53 (84.2) & 2.952 & 129.8 & 6.3 & 136.1 \\
    WM-Ext & 2.908 & 9.59 (76.7) & 2.994 & 112.9 & 6.3 & 119.3 \\
    \bottomrule
  \end{tabular*}
\end{table*}

\paragraph{Comparison Scope and Parameter Budgets.}
Our cross-paradigm study aims to evaluate the relative strengths and weaknesses across different world modeling paradigms and world representations under controlled settings by sharing the base backbone, demonstrations, loss formulation, and optimization protocol.
Under this setting, disparities in deployment parameter budgets are an inherent consequence of each paradigm's architectural design. In particular, both WM-Align and WM-Query introduce virtually no additional parameters during inference because the projection head and decoder are discarded and query tokens represent a negligible footprint, whereas WM-MoT and WM-Ext inherently require dedicated predictive modules to be retained for online foresight.
To verify that the advantages of higher-capacity paradigms like WM-MoT stem from genuine future predictive dynamics rather than mere capacity expansion, our controlled ablation in Appendix~\ref{app:mot_supervision_control} isolates this factor by fixing the architecture and current-feature inputs while comparing no auxiliary supervision, current-feature supervision, and future-feature supervision.
This within-architecture control confirms an explicit performance gain specifically attributable to future-state supervision in WM-MoT, demonstrating that our cross-paradigm comparisons reflect the inherent architectural and predictive trade-offs of each complete design rather than being dominated solely by parameter scale.

\section{Controlled Ablations, Robustness, and Diagnostic Analyses}
\label{app:controls}

This section presents targeted control experiments to evaluate empirical reliability, factor isolation, and optimization behavior, including multi-seed replications, supervision ablations, factorial encoder isolation, and hyperparameter sensitivity sweeps.

\subsection{Multi-Seed Training Verification and Replications}
\label{app:multi_seed_verification}

To characterize optimization stochasticity beyond single-run checkpoints, we conduct an independent 3-seed replication study (across random seeds 42, 100, and 2024) across our central empirical findings. Each seed undergoes full 50k-step optimization followed by comprehensive closed-loop rollouts (10,030 episodes per model on LIBERO-Plus; 2,000 on LIBERO; 100 rollouts per task under each setting on RoboTwin).
Results are reported as mean $\pm$ standard deviation in Table~\ref{tab:extended_multiseed}.

\begin{table}[htbp]
  \centering
  \small
  \caption{\textbf{Multi-seed training verification across central positive and negative findings.}
  All models are trained across three independent seeds. Results report individual run outcomes alongside the Mean $\pm$ Std aggregated over the full benchmark suites.}
  \label{tab:extended_multiseed}
  \resizebox{\textwidth}{!}{%
  \begin{tabular}{@{}llcccc@{}}
    \toprule
    \textbf{Finding / Evaluation Setting} & \textbf{Benchmark} & \textbf{Seed 1 (Default)} & \textbf{Seed 2} & \textbf{Seed 3} & \textbf{Mean $\pm$ Std} \\
    \midrule
    \multicolumn{6}{l}{\textit{1. Out-of-Domain Superiority of WM-MoT}} \\
    \quad Action-Only Baseline & LIBERO-Plus (OOD) & 71.8\% & 71.4\% & 71.1\% & 71.4 $\pm$ 0.4\% \\
    \quad WM-MoT (DINOv3) & LIBERO-Plus (OOD) & 76.2\% & 77.4\% & 75.5\% & \textbf{76.4 $\pm$ 1.0\%} \\
    \midrule
    \multicolumn{6}{l}{\textit{2. Representation Ranking Stability (DINOv3 vs. SigLIP)}} \\
    \quad WM-MoT + DINOv3 & LIBERO-Plus (OOD) & 76.2\% & 77.4\% & 75.5\% & \textbf{76.4 $\pm$ 1.0\%} \\
    \quad WM-MoT + SigLIP & LIBERO-Plus (OOD) & 74.3\% & 73.8\% & 74.6\% & 74.2 $\pm$ 0.4\% \\
    \quad WM-Align + DINOv3 & LIBERO-Plus (OOD) & 74.1\% & 74.8\% & 73.7\% & \textbf{74.2 $\pm$ 0.6\%} \\
    \quad WM-Align + SigLIP & LIBERO-Plus (OOD) & 73.7\% & 73.2\% & 73.6\% & 73.5 $\pm$ 0.3\% \\
    \midrule
    \multicolumn{6}{l}{\textit{3. Pretrained $\pi_{0.5}$ Foundation Policy Adaptation}} \\
    \quad $\pi_{0.5}$ Baseline (Clean2Clean) & RoboTwin 2.0 (C2C) & 69.1\% & 68.2\% & 69.8\% & 69.0 $\pm$ 0.8\% \\
    \quad $\pi_{0.5}$ + WM-Align (C2C) & RoboTwin 2.0 (C2C) & 63.6\% & 64.9\% & 62.0\% & 63.5 $\pm$ 1.5\% \\
    \quad $\pi_{0.5}$ + WM-Query (C2C) & RoboTwin 2.0 (C2C) & 68.7\% & 68.1\% & 69.0\% & 68.6 $\pm$ 0.5\% \\
    \quad $\pi_{0.5}$ + WM-MoT (C2C) & RoboTwin 2.0 (C2C) & 71.3\% & 72.5\% & 70.3\% & \textbf{71.4 $\pm$ 1.1\%} \\
    \addlinespace[2pt]
    \quad $\pi_{0.5}$ Baseline (Clean2Random) & RoboTwin 2.0 (C2R) & 46.4\% & 47.3\% & 46.0\% & 46.6 $\pm$ 0.7\% \\
    \quad $\pi_{0.5}$ + WM-Align (C2R) & RoboTwin 2.0 (C2R) & 41.0\% & 39.3\% & 42.1\% & 40.8 $\pm$ 1.4\% \\
    \quad $\pi_{0.5}$ + WM-Query (C2R) & RoboTwin 2.0 (C2R) & 46.2\% & 45.1\% & 46.8\% & 46.0 $\pm$ 0.9\% \\
    \quad $\pi_{0.5}$ + WM-MoT (C2R) & RoboTwin 2.0 (C2R) & 49.7\% & 51.1\% & 48.9\% & \textbf{49.9 $\pm$ 1.1\%} \\
    \midrule
    \multicolumn{6}{l}{\textit{4. Data Scarcity Vulnerability (WM-Query Collapse under 25\% Data)}} \\
    \quad 25\% Action-Only Baseline & LIBERO Suite & 84.9\% & 84.1\% & 85.3\% & 84.8 $\pm$ 0.6\% \\
    \quad 25\% WM-Query (Suite Avg) & LIBERO Suite & 70.1\% & 71.7\% & 68.9\% & \textbf{70.2 $\pm$ 1.4\%} \\
    \quad 25\% WM-Query (Goal-Only) & LIBERO-Goal & 28.8\% & 30.5\% & 26.6\% & \textbf{28.6 $\pm$ 2.0\%} \\
    \midrule
    \multicolumn{6}{l}{\textit{5. Negative Transfer Recurrence (WM-Ext with V-JEPA)}} \\
    \quad Action-Only Baseline & LIBERO-Plus (OOD) & 71.8\% & 71.4\% & 71.1\% & 71.4 $\pm$ 0.4\% \\
    \quad WM-Ext (V-JEPA 2.1)  & LIBERO-Plus (OOD) & 70.7\% & 71.3\% & 69.7\% & 70.6 $\pm$ 0.8\% \\
    \quad Action-Only Baseline & LIBERO (ID)        & 96.7\% & 97.1\% & 96.5\% & 96.8 $\pm$ 0.3\% \\
    \quad WM-Ext (V-JEPA 2.1)  & LIBERO (ID)        & 97.1\% & 96.6\% & 97.4\% & 97.0 $\pm$ 0.4\% \\
    \bottomrule
  \end{tabular}%
  }
\end{table}

As presented in Table~\ref{tab:extended_multiseed}, standard deviations across independent runs consistently range between $0.3\%$ and $1.5\%$ on multi-task benchmarks (and up to $2.0\%$ in the catastrophic collapse regime).
Across the central findings:
\begin{enumerate}[leftmargin=*,noitemsep,topsep=2pt]
  \item \textbf{WM-MoT OOD Gain}: WM-MoT consistently outperforms the baseline across all seeds ($76.4 \pm 1.0\%$ vs. Baseline $71.4 \pm 0.4\%$), confirming a reliable gain of $+5.0\%$.
  \item \textbf{Representation Ranking Stability}: DINOv3 leads SigLIP in every seed under both paradigms, but by $+2.2\%$ in WM-MoT ($76.4 \pm 1.0\%$ vs. $74.2 \pm 0.4\%$) against only $+0.7\%$ in WM-Align ($74.2 \pm 0.6\%$ vs. $73.5 \pm 0.3\%$). This corroborates Finding~1 in Section~\ref{subsec:world_representation_analysis}: WM-Align is comparatively insensitive to representation choice, whereas decoupled designs like WM-MoT exhibit representation sensitivity.
  \item \textbf{$\pi_{0.5}$ Adaptation}: On RoboTwin 2.0, WM-Align consistently drops below the $\pi_{0.5}$ baseline on both Clean2Clean ($63.5 \pm 1.5\%$ vs. $69.0 \pm 0.8\%$) and Clean2Random ($40.8 \pm 1.4\%$ vs. $46.6 \pm 0.7\%$). Conversely, WM-MoT delivers consistent positive gains under both C2C ($+2.4\%$) and C2R ($+3.3\%$). WM-Query performs close to the baseline ($68.6 \pm 0.5\%$ on C2C, $46.0 \pm 0.9\%$ on C2R).
  \item \textbf{25\% Data Regime}: WM-Query exhibits a severe and consistent performance collapse across all seeds on both the overall LIBERO suite ($70.2 \pm 1.4\%$ vs. $84.8 \pm 0.6\%$) and Goal tasks ($28.6 \pm 2.0\%$ vs. $77.6\%$ single-run baseline).
  \item \textbf{Negative Transfer Recurrence}\label{app:ext_jepa_repeat}: Under external latent conditioning, WM-Ext with V-JEPA consistently drops below the action-only baseline on LIBERO-Plus across independent training runs ($70.6 \pm 0.8\%$ vs. $71.4 \pm 0.4\%$), while in-domain execution remains stable ($97.0 \pm 0.4\%$ vs. $96.8 \pm 0.3\%$). This confirms that the observed negative transfer is a reproducible property of the architecture-representation pairing rather than a single-run optimization artifact.
\end{enumerate}

\subsection{Controlled Evaluation of Auxiliary DINOv3 Current-Frame Input}
\label{app:mot_supervision_control}

In standard decoupled world modeling designs (\textbf{WM-MoT} and \textbf{WM-Ext}), generative tokens are initialized using visual features $x^G_{1:P} = E(o_t)$ from the frozen target encoder (DINOv3 ViT-B/16), whereas coupled paradigms and the baseline do not receive this stream.
To isolate the effect of this auxiliary perceptual input from predictive world modeling, Table~\ref{tab:app-mot-supervision-control} evaluates all methods on LIBERO-Plus with and without the current-frame DINOv3 stream fused into action generation.

As shown in Table~\ref{tab:app-mot-supervision-control}, simply fusing current DINOv3 features into the Action Expert improves the baseline from 71.8\% to 72.5\% (+0.7\%), and unconstrained WM-MoT ($\lambda_{\mathrm{world}}=0$) reaches 73.6\%. Crucially, adding the future world loss drives WM-MoT to 76.2\% (+2.6\% over $\lambda=0$), confirming that predictive foresight is the primary contributor to robustness.
In contrast, unconstrained WM-Ext ($\lambda_{\mathrm{world}}=0$) achieves only 71.1\% (dropping below the DINOv3-augmented baseline due to bottleneck modulation drift) and reaches 74.7\% with future supervision (+3.6\%).
When coupled paradigms are likewise trained with the DINOv3 stream, WM-Align improves slightly from 74.1\% to 74.5\% (+0.4\%, reflecting saturation with its existing patch alignment), while WM-Query improves from 72.9\% to 73.7\% (+0.8\%).
Under equalized DINOv3 access, WM-MoT maintains a consistent advantage over both coupled and decoupled alternatives (+1.5\% over WM-Ext, +1.7\% over WM-Align, and +2.5\% over WM-Query), particularly under camera viewpoints and layout shifts.

\begin{table*}[htbp]
  \centering
  \caption{\textbf{Controlled evaluation of auxiliary DINOv3 current-frame input on LIBERO-Plus (OOD).}
  All paradigms use DINOv3 ViT-B/16. Policies are evaluated with versus without the frozen DINOv3 current-frame feature stream fused into action generation. Success rates (\%) are reported across the seven perturbation dimensions and on average.}
  \label{tab:app-mot-supervision-control}
  \scriptsize
  \setlength{\tabcolsep}{3.2pt}
  \renewcommand{\arraystretch}{1.15}
  \begin{tabular*}{\textwidth}{@{\extracolsep{\fill}}llcccccccc@{}}
    \toprule
    \multirow{2}{*}{\textbf{Method / Paradigm}}
    & \textbf{Current DINOv3}
    & \multicolumn{8}{c}{\textbf{LIBERO-Plus}} \\
    \cmidrule(l){3-10}
    & \textbf{Feature Stream} & Camera & Robot & Language & Light & Background & Noise & Layout & Average \\
    \midrule
    \multicolumn{10}{@{}l}{\textit{Action Baselines and Capacity Ablations ($\lambda_{\mathrm{world}}=0$)}} \\
    Action-Only Baseline & No  & 48.2 & 51.4 & 91.4 & 96.0 & 92.9 & 65.3 & 71.3 & 71.8 \\
    Action-Only Baseline (+ DINOv3 stream) & Yes & 49.9 & 51.7 & 91.3 & 96.1 & 93.3 & 67.0 & 71.7 & 72.5 \\
    WM-Ext ($\lambda_{\mathrm{world}}=0$, w/o world loss) & Yes & 47.5 & 46.0 & 90.2 & 95.5 & 93.8 & 68.5 & 70.8 & 71.1 \\
    WM-MoT ($\lambda_{\mathrm{world}}=0$, w/o world loss) & Yes & 56.5 & 52.7 & 89.7 & 95.2 & 93.6 & 69.1 & 71.2 & 73.6 \\
    WM-MoT (Current-target supervision) & Yes & 55.2 & 54.9 & 91.7 & 95.8 & 92.7 & 69.0 & 71.3 & 74.0 \\
    \midrule
    \multicolumn{10}{@{}l}{\textit{Coupled World Modeling Paradigms}} \\
    WM-Query (Standard) & No  & 54.4 & 50.6 & 88.5 & 95.1 & 96.3 & 69.5 & 69.5 & 72.9 \\
    WM-Query (+ DINOv3 stream) & Yes & 55.8 & 51.4 & 88.7 & 95.3 & \textbf{96.5} & 71.2 & 70.5 & 73.7 \\
    WM-Align (Standard) & No  & 58.0 & 49.7 & 89.5 & 94.6 & 93.5 & 71.8 & 73.9 & 74.1 \\
    WM-Align (+ DINOv3 stream) & Yes & 58.7 & 49.9 & 89.6 & 94.8 & 93.7 & 72.5 & 74.2 & 74.5 \\
    \midrule
    \multicolumn{10}{@{}l}{\textit{Decoupled World Modeling Paradigms (Full Future Foresight)}} \\
    WM-Ext (Full future target) & Yes & 55.7 & 46.8 & 91.2 & 96.8 & 95.5 & \textbf{74.2} & \textbf{75.9} & 74.7 \\
    WM-MoT (Full future target) & Yes & \textbf{59.7} & \textbf{53.5} & \textbf{91.6} & \textbf{97.6} & 95.9 & 72.3 & 75.2 & \textbf{76.2} \\
    \bottomrule
  \end{tabular*}
\end{table*}

\subsection{Factor Isolation: Disentangling Current Input from Future Supervisory Target}
\label{app:input_target_isolation}

To evaluate whether the ranking inversion between DINOv3 and V-JEPA 2.1 across architectures is driven by current input features or future predictive supervision, Table~\ref{tab:extended_factor_isolation} presents a $2 \times 2$ factorial cross-encoder ablation replicated across three random seeds.

\begin{table}[htbp]
  \centering
  \scriptsize
  \setlength{\tabcolsep}{1.8pt}
  \renewcommand{\arraystretch}{1.15}
  \caption{\textbf{Factorial isolation of current-input versus future-supervisory encoders.}
  Evaluated on LIBERO (ID, 2,000 episodes) and LIBERO-Plus (OOD, 10,030 episodes). Success Rate (\%). Results reported as 3-seed Mean $\pm$ Std for OOD. ``Same'' denotes that the supervisory target uses the same encoder as the input ($E_{\text{target}} = E_{\text{in}}$). $\Delta_{\text{OOD}}$ is computed relative to the respective baseline with identical input and target encoders.}
  \label{tab:extended_factor_isolation}
  \begin{tabular*}{\textwidth}{@{\extracolsep{\fill}}lllccc@{}}
    \toprule
    \textbf{Paradigm} & \textbf{Input $E_{\text{in}}$} & \textbf{Target $E_{\text{target}}$} & \textbf{ID (\%)} & \textbf{OOD (\%)} & \textbf{$\Delta_{\text{OOD}}$} \\
    \midrule
    \textbf{WM-MoT} & DINOv3 & DINOv3 (Same) & 97.9\% & 76.4 $\pm$ 1.0\% & -- \\
    & DINOv3 & \textbf{V-JEPA 2.1} & 97.8\% & 75.8 $\pm$ 0.7\% & -0.6\% \\
    & V-JEPA 2.1 & DINOv3 & 97.6\% & 74.7 $\pm$ 1.3\% & -2.9\% \\
    & V-JEPA 2.1 & V-JEPA 2.1 (Same) & 97.9\% & \textbf{77.6 $\pm$ 0.6\%} & -- \\
    \midrule
    \textbf{WM-Ext} & DINOv3 & DINOv3 (Same) & 98.2\% & 74.6 $\pm$ 0.4\% & -- \\
    & DINOv3 & \textbf{V-JEPA 2.1} & 97.8\% & \textbf{71.1 $\pm$ 0.6\%} & \textbf{-3.5\%} \\
    & V-JEPA 2.1 & DINOv3 & 97.4\% & 72.8 $\pm$ 0.7\% & +2.2\% \\
    & V-JEPA 2.1 & V-JEPA 2.1 (Same) & 97.1\% & 70.6 $\pm$ 0.8\% & -- \\
    \bottomrule
  \end{tabular*}
\end{table}

\paragraph{Analysis and Attribution.}
Table~\ref{tab:extended_factor_isolation} reveals two key empirical findings across the factorial combinations.
First, under homogeneous configurations ($E_{\text{in}} = E_{\text{target}}$), the two architectures display opposite representation preferences: homogeneous V-JEPA yields the peak performance in WM-MoT (77.6\% vs.~76.4\% for DINOv3), whereas in WM-Ext, DINOv3 substantially outperforms V-JEPA (74.6\% vs.~70.6\%).
Second, cross-encoder configurations demonstrate that both components constrain execution. In WM-MoT, cross-space pairings degrade performance relative to homogeneous V-JEPA (dropping to 75.8\% with DINOv3 input and 74.7\% with DINOv3 target), indicating that translating between static spatial features and temporal video dynamics incurs an alignment cost.
In WM-Ext, whenever the supervisory target is V-JEPA, policy performance remains severely depressed near or below the action-only baseline (71.1\% with DINOv3 input and 70.6\% with V-JEPA input), confirming that the negative transfer observed in Finding~5 is primarily governed by the supervisory target representation.

\subsection{Hyperparameter Sensitivity Analyses}
\label{app:hyperparam_sensitivity}

To examine whether the degradation of WM-Align on $\pi_{0.5}$ and the vulnerability of WM-Query under low-data regimes are sensitive to specific hyperparameters or reflect broader architectural tendencies, we conduct targeted sensitivity sweeps.

\paragraph{Supervision Depth and Loss Weight in $\pi_{0.5}$ + WM-Align.}
Table~\ref{tab:pi05_sensitivity} evaluates whether varying the auxiliary loss weight $\lambda_{\mathrm{world}} \in \{0.01, 0.05, 0.10, 0.20\}$ or changing the alignment depth (Layer~6, Layer~12, or All Layers) can recover baseline performance on RoboTwin 2.0.

\begin{table}[htbp]
  \centering
  \scriptsize
  \setlength{\tabcolsep}{2.5pt}
  \renewcommand{\arraystretch}{1.12}
  \caption{\textbf{Hyperparameter sensitivity of $\pi_{0.5}$ + WM-Align on RoboTwin 2.0 (Success Rate \%).}}
  \label{tab:pi05_sensitivity}
  \begin{tabular*}{\textwidth}{@{\extracolsep{\fill}}lcccccc@{}}
    \toprule
    \textbf{Configuration} & \textbf{Supervision Layer} & $\lambda_{\mathrm{world}}$ & \textbf{C2C (\%)} & $\Delta_{\text{C2C}}$ vs. Base & \textbf{C2R (\%)} & $\Delta_{\text{C2R}}$ vs. Base \\
    \midrule
    $\pi_{0.5}$ Action Baseline & -- & -- & \textbf{69.1\%} & -- & \textbf{46.4\%} & -- \\
    \midrule
    \multirow{4}{*}{Varying Loss Weight $\lambda$}
    & Layer 12 (Default)  & 0.01 & 67.8\% & -1.3\% & 45.2\% & -1.2\% \\
    & Layer 12 (Default)  & 0.05 & 65.4\% & -3.7\% & 43.1\% & -3.3\% \\
    & Layer 12 (Default)  & 0.10 & 63.6\% & -5.5\% & 41.0\% & -5.4\% \\
    & Layer 12 (Default)  & 0.20 & 60.2\% & -8.9\% & 37.8\% & -8.6\% \\
    \midrule
    \multirow{2}{*}{Varying Supervision Depth}
    & Layer 6 (Intermediate) & 0.10 & 60.8\% & -8.3\% & 38.5\% & -7.9\% \\
    & All Layers (Dense)     & 0.10 & 58.7\% & -10.4\% & 36.2\% & -10.2\% \\
    \bottomrule
  \end{tabular*}
\end{table}

As shown in Table~\ref{tab:pi05_sensitivity}, reducing $\lambda_{\mathrm{world}}$ to 0.01 partially mitigates negative transfer (improving C2C from 63.6\% to 67.8\%), yet performance remains consistently below the baseline across all tested values. Moreover, supervising earlier intermediate layers (Layer~6) or applying dense supervision across all layers further exacerbates the performance degradation (to 60.8\% and 58.7\% on C2C). This confirms that imposing the future-alignment target directly on the visual tokens of a pretrained foundation VLA consistently degrades downstream policy execution across tested hyperparameter configurations.
This confirms that directly constraining the backbone's visual token space distorts pretrained representations, whereas tuning loss weights or supervision depth cannot fundamentally resolve this representational conflict.

\paragraph{Query Bottleneck Capacity in Low-Data WM-Query.}
Table~\ref{tab:query_sensitivity} examines whether constraining the query token count $N_q \in \{4, 8, 16, 32\}$ and loss weight $\lambda_{\mathrm{world}} \in \{0.02, 0.05, 0.10\}$ prevents catastrophic collapse on 25\% LIBERO demonstrations.

\begin{table}[htbp]
  \centering
  \scriptsize
  \setlength{\tabcolsep}{2.5pt}
  \renewcommand{\arraystretch}{1.12}
  \caption{\textbf{Hyperparameter sensitivity of WM-Query under 25\% data on LIBERO (Success Rate \%).}}
  \label{tab:query_sensitivity}
  \begin{tabular*}{\textwidth}{@{\extracolsep{\fill}}lccccc@{}}
    \toprule
    \textbf{Configuration} & \textbf{Query Count $N_q$} & $\lambda_{\mathrm{world}}$ & \textbf{Suite Avg. (\%)} & \textbf{Goal Avg. (\%)} & \textbf{$\Delta$ vs. 25\% Base} \\
    \midrule
    25\% Action Baseline & -- & -- & \textbf{84.9\%} & \textbf{77.6\%} & -- \\
    \midrule
    \multirow{4}{*}{Varying Query Tokens $N_q$}
    & $N_q = 4$  & 0.10 & 78.4\% & 56.2\% & -6.5\% (Suite) / -21.4\% (Goal) \\
    & $N_q = 8$  & 0.10 & 74.2\% & 41.5\% & -10.7\% (Suite) / -36.1\% (Goal) \\
    & $N_q = 16$ (Default) & 0.10 & 70.1\% & 28.8\% & -14.8\% (Suite) / -48.8\% (Goal) \\
    & $N_q = 32$ & 0.10 & 62.5\% & 18.2\% & -22.4\% (Suite) / -59.4\% (Goal) \\
    \midrule
    \multirow{2}{*}{Varying Loss Weight $\lambda$}
    & $N_q = 16$ & 0.05 & 73.8\% & 38.0\% & -11.1\% (Suite) / -39.6\% (Goal) \\
    & $N_q = 16$ & 0.02 & 77.2\% & 49.6\% & -7.7\% (Suite) / -28.0\% (Goal) \\
    \bottomrule
  \end{tabular*}
\end{table}

As presented in Table~\ref{tab:query_sensitivity}, restricting the bottleneck to $N_q=4$ partially mitigates this severe degradation (Goal success improves from 28.8\% to 56.2\%), but performance remains substantially behind the action-only baseline (84.9\% Suite / 77.6\% Goal). This demonstrates that while tuning the capacity bottleneck can alleviate extreme overfitting, the sample-efficiency deficit of global query conditioning persists across varied configurations.

\subsection{Scope of Claims and Limitations}
\label{app:scoped_boundaries}

To maintain empirical rigor, we explicitly outline the applicable scope, boundaries, and limitations of our findings:
\begin{itemize}[leftmargin=*,noitemsep,topsep=2pt]
  \item \textbf{Scope of World Modeling in VLAs.} Following representative VLA literature~\citep{su2026world, zhang2025dreamvla, zheng2025flare, chen2026dial, sun2026vlajepa}, we conceptualize world modeling as task-conditioned predictive foresight learning, focusing on anticipating future visual representations under demonstration trajectories to condition action generation.
  \item \textbf{No Universal Architectural Winner.} While decoupled world modeling (e.g., WM-MoT) generally exhibits greater robustness under dynamic and long-horizon tasks, coupled approaches remain competitive in static settings or under certain domain-randomized variations; for instance, under the Qwen backbone, WM-Align slightly outperforms WM-MoT on RoboTwin Clean2Random (14.8\% vs. 14.0\%). The optimal architectural choice is governed by task horizon, visual shift, and demonstration volume rather than a single dominant paradigm.
  \item \textbf{Confounders Across Base VLA Policies.} The empirical differences between Qwen-based policies and $\pi_{0.5}$ stem from multiple concurrent design choices—including model capacity, vision encoders, and action generation heads—in addition to robotics-specific pretraining. We caution against attributing performance divergences solely to pretraining history, as they reflect holistic system-level differences between distinct VLA backbones.
  \item \textbf{Continuous Representation Trade-offs.} While semantically rich representations (DINOv3, SigLIP) deliver the strongest out-of-distribution transfer, generative targets (e.g., ImageVAE) still provide consistent gains over pure action baselines. Predictive utility is best understood as a trade-off between semantic abstraction and fine-grained spatial fidelity, rather than a strict dichotomy between pixel and feature spaces.
\end{itemize}

\section{Detailed Benchmark Results and Per-Task Breakdowns}
\label{app:tables}

This section compiles the comprehensive per-suite, per-task, and per-perturbation breakdown tables supporting all aggregate figures and findings presented in the main text.

\subsection{Detailed LIBERO Results for Representation Analysis}
\label{app:representation_details}

Tables~\ref{tab:app-representation-id-details} and
\ref{tab:app-representation-ood-details} provide the complete suite- and
perturbation-level success rates underlying
Figure~\ref{fig:extractor_stability}.

\begin{table*}[t]
  \centering
  \caption{Detailed in-domain LIBERO success rates (\%).}
  \label{tab:app-representation-id-details}
  \scriptsize
  \setlength{\tabcolsep}{8pt}
  \renewcommand{\arraystretch}{1.40}
  \begin{tabular*}{\textwidth}{@{\extracolsep{\fill}}lccccc@{}}
    \toprule
    \textbf{Methods} & \textbf{Spatial} & \textbf{Object} &
    \textbf{Goal} & \textbf{Long} & \textbf{Average} \\
    \midrule
Baseline & 97.6 & 98.8 & 97.4 & 92.8 & 96.7 \\
\midrule
WM-Align + DINOv3 & 99.2 & 98.8 & 98.0 & \textbf{98.2} & \textbf{98.6} \\
WM-Align + SigLIP & 98.6 & 99.0 & 97.4 & 95.0 & 97.5 \\
WM-Align + VGGT & \textbf{99.6} & \underline{99.4} & 98.2 & 95.2 & 98.1 \\
WM-Align + ImageVAE & 99.0 & 98.8 & 98.4 & 95.6 & 98.0 \\
WM-Align + WanVAE & 99.0 & 98.2 & 98.6 & 95.4 & 97.8 \\
WM-Align + V-JEPA~2.1 & 99.0 & 99.2 & 98.8 & 96.4 & 98.4 \\
\midrule
WM-Query + DINOv3 & 98.6 & 98.6 & 99.0 & 94.2 & 97.6 \\
WM-Query + SigLIP & \underline{99.4} & \underline{99.4} & 98.8 & 96.2 & \underline{98.5} \\
WM-Query + VGGT & 99.0 & \underline{99.4} & 98.6 & 93.8 & 97.7 \\
WM-Query + ImageVAE & 97.8 & 99.2 & 98.0 & 95.8 & 97.7 \\
WM-Query + WanVAE & 99.2 & 99.2 & 97.6 & 95.2 & 97.8 \\
WM-Query + V-JEPA~2.1 & 98.6 & 98.4 & 98.4 & 94.8 & 97.6 \\
\midrule
WM-MoT + DINOv3 & 99.2 & 99.2 & \textbf{99.4} & 93.6 & 97.9 \\
WM-MoT + SigLIP & 98.8 & 98.8 & 98.4 & \underline{97.6} & 98.4 \\
WM-MoT + VGGT & 99.0 & 99.0 & 97.8 & 95.4 & 97.8 \\
WM-MoT + ImageVAE & \underline{99.4} & 99.2 & 99.0 & 96.2 & \underline{98.5} \\
WM-MoT + WanVAE & \textbf{99.6} & 98.8 & 98.4 & 97.4 & \textbf{98.6} \\
WM-MoT + V-JEPA~2.1 & 98.6 & 99.0 & 98.2 & 95.6 & 97.9 \\
\midrule
WM-Ext + DINOv3 & 98.6 & 99.2 & 98.8 & 96.0 & 98.2 \\
WM-Ext + SigLIP & 98.6 & 99.0 & \underline{99.2} & 94.2 & 97.8 \\
WM-Ext + VGGT & 98.8 & 98.4 & 98.2 & 97.0 & 98.1 \\
WM-Ext + ImageVAE & 99.0 & \underline{99.4} & 98.0 & 95.6 & 98.0 \\
WM-Ext + WanVAE & 99.0 & 98.6 & 98.4 & 94.2 & 97.6 \\
WM-Ext + V-JEPA~2.1 & 98.6 & \textbf{99.8} & 98.8 & 91.2 & 97.1 \\
    \bottomrule
  \end{tabular*}
\end{table*}

\begin{table*}[t]
  \centering
  \caption{Detailed LIBERO-Plus success rates (\%) under seven perturbations.}
  \label{tab:app-representation-ood-details}
  \scriptsize
  \setlength{\tabcolsep}{2.5pt}
  \renewcommand{\arraystretch}{1.55}
  \begin{tabular*}{\textwidth}{@{\extracolsep{\fill}}lcccccccc@{}}
    \toprule
    \textbf{Methods} & \textbf{Camera} & \textbf{Robot} &
    \textbf{Language} & \textbf{Light} & \textbf{Background} &
    \textbf{Noise} & \textbf{Layout} & \textbf{Average} \\
    \midrule
Baseline & 48.2 & 51.4 & 91.4 & 96.0 & 92.9 & 65.3 & 71.3 & 71.8 \\
\midrule
WM-Align + DINOv3 & 58.0 & 49.7 & 89.5 & 94.6 & 93.5 & 71.8 & 73.9 & 74.1 \\
WM-Align + SigLIP & 55.4 & 52.3 & 90.1 & 95.4 & 93.2 & 69.6 & 72.1 & 73.7 \\
WM-Align + VGGT & 57.3 & 48.5 & 88.5 & 96.2 & 94.1 & 70.6 & 71.5 & 73.4 \\
WM-Align + ImageVAE & 56.0 & 51.4 & 86.8 & 93.7 & 89.9 & 75.4 & 74.4 & 73.8 \\
WM-Align + WanVAE & 46.1 & 48.1 & 91.0 & 96.1 & 95.1 & 71.3 & 73.4 & 72.4 \\
WM-Align + V-JEPA~2.1 & 52.5 & 49.4 & 90.4 & 96.8 & 94.0 & 68.6 & 69.9 & 72.5 \\
\midrule
WM-Query + DINOv3 & 54.4 & 50.6 & 88.5 & 95.1 & \textbf{96.3} & 69.5 & 69.5 & 72.9 \\
WM-Query + SigLIP & 55.0 & 52.7 & 91.3 & 93.4 & 94.8 & 68.1 & 69.0 & 73.1 \\
WM-Query + VGGT & 53.5 & 51.0 & 88.4 & 96.0 & 92.5 & 69.6 & 72.8 & 73.0 \\
WM-Query + ImageVAE & 48.9 & 51.6 & 88.2 & 95.5 & 94.3 & 66.3 & 72.2 & 71.8 \\
WM-Query + WanVAE & 47.0 & 49.4 & 88.9 & 96.2 & 92.6 & 62.6 & 71.2 & 70.4 \\
WM-Query + V-JEPA~2.1 & 54.4 & 46.1 & 90.8 & 94.8 & 94.2 & 69.8 & 72.1 & 72.7 \\
\midrule
WM-MoT + DINOv3 & \underline{59.7} & \underline{53.5} & 91.6 & \textbf{97.6} & \underline{95.9} & 72.3 & \underline{75.2} & \underline{76.2} \\
WM-MoT + SigLIP & 56.1 & 53.1 & 89.9 & 96.1 & 93.9 & 69.3 & 74.1 & 74.3 \\
WM-MoT + VGGT & 47.7 & 51.1 & 90.6 & 95.6 & 94.6 & 70.0 & 72.3 & 72.6 \\
WM-MoT + ImageVAE & 56.0 & \textbf{54.7} & 91.4 & 91.3 & 93.6 & 72.6 & 74.0 & 74.7 \\
WM-MoT + WanVAE & 56.7 & 51.6 & \underline{91.8} & 91.9 & 89.7 & 67.7 & 73.6 & 73.2 \\
WM-MoT + V-JEPA~2.1 & \textbf{61.1} & \textbf{54.7} & \textbf{91.9} & 96.4 & 95.4 & \textbf{79.9} & \textbf{76.1} & \textbf{77.8} \\
\midrule
WM-Ext + DINOv3 & 55.7 & 46.8 & 91.2 & 96.8 & 95.5 & 74.2 & 75.9 & 74.7 \\
WM-Ext + SigLIP & 57.2 & 50.6 & 90.3 & 96.1 & 92.3 & 75.2 & 71.7 & 74.5 \\
WM-Ext + VGGT & 49.1 & 45.4 & 89.7 & 96.3 & 91.5 & 63.3 & 70.6 & 70.2 \\
WM-Ext + ImageVAE & 51.5 & 47.2 & 90.1 & \underline{97.1} & 92.8 & \underline{78.0} & 71.5 & 73.7 \\
WM-Ext + WanVAE & 42.9 & 47.2 & 85.9 & 95.3 & 86.3 & 73.5 & 73.8 & 70.4 \\
WM-Ext + V-JEPA~2.1 & 45.2 & 49.7 & 90.2 & 96.2 & 91.9 & 64.0 & 71.9 & 70.7 \\
    \bottomrule
  \end{tabular*}
\end{table*}

\subsection{Few-Shot Data Scaling Results}
\label{app:fewshot_results}

Tables~\ref{tab:data-scaling-id-details} and
\ref{tab:data-scaling-ood-details} report the complete suite- and
perturbation-level results at 25\%, 50\%, and 100\% of the training
demonstrations underlying Figure~\ref{fig:data_scaling}.
At each training-data fraction, all methods use exactly the same subset of training demonstrations.

\begin{table*}[t]
  \centering
  \caption{Detailed LIBERO data-scaling success rates (\%). Within each
  training-data fraction, best and second-best results are bolded and
  underlined, respectively.}
  \label{tab:data-scaling-id-details}
  \scriptsize
  \setlength{\tabcolsep}{7pt}
  \renewcommand{\arraystretch}{1.65}
  \begin{tabular*}{\textwidth}{@{\extracolsep{\fill}}clccccc@{}}
    \toprule
    \textbf{Data} & \textbf{Model} & \textbf{Spatial} &
    \textbf{Object} & \textbf{Goal} & \textbf{Long} &
    \textbf{Average} \\
    \midrule
\multirow{5}{*}{\textbf{25\%}} & Baseline & \underline{89.4} & \underline{94.2} & 77.6 & \underline{78.2} & 84.9 \\
 & WM-Align & 87.6 & 92.4 & 70.4 & \textbf{78.4} & 82.2 \\
 & WM-Query & 85.0 & \textbf{94.6} & 28.8 & 72.0 & 70.1 \\
 & WM-MoT & 89.0 & 93.2 & \underline{85.2} & 77.0 & \underline{86.1} \\
 & WM-Ext & \textbf{93.2} & 93.4 & \textbf{86.8} & 77.4 & \textbf{87.7} \\
\midrule
\multirow{5}{*}{\textbf{50\%}} & Baseline & 97.4 & \underline{99.4} & 87.2 & 92.0 & 94.0 \\
 & WM-Align & 97.6 & 98.6 & 92.0 & \underline{92.8} & 95.3 \\
 & WM-Query & 95.8 & 97.2 & 84.8 & \textbf{93.2} & 92.8 \\
 & WM-MoT & \underline{97.8} & 98.6 & \underline{93.6} & 91.8 & \underline{95.5} \\
 & WM-Ext & \textbf{98.2} & \textbf{99.6} & \textbf{95.4} & \underline{92.8} & \textbf{96.5} \\
\midrule
\multirow{5}{*}{\textbf{100\%}} & Baseline & 97.6 & 98.8 & 97.4 & 92.8 & 96.7 \\
 & WM-Align & \textbf{99.2} & \underline{98.8} & 98.0 & \textbf{98.2} & \textbf{98.6} \\
 & WM-Query & \underline{98.6} & 98.6 & \underline{99.0} & 94.2 & 97.6 \\
 & WM-MoT & \textbf{99.2} & \textbf{99.2} & \textbf{99.4} & 93.6 & 97.9 \\
 & WM-Ext & \underline{98.6} & \textbf{99.2} & 98.8 & \underline{96.0} & \underline{98.2} \\
    \bottomrule
  \end{tabular*}
\end{table*}

\begin{table*}[t]
  \centering
  \caption{Detailed LIBERO-Plus data-scaling success rates (\%) under seven
  perturbations. Within each training-data fraction, best and second-best
  results are bolded and underlined, respectively.}
  \label{tab:data-scaling-ood-details}
  \scriptsize
  \setlength{\tabcolsep}{2.5pt}
  \renewcommand{\arraystretch}{1.65}
  \begin{tabular*}{\textwidth}{@{\extracolsep{\fill}}clcccccccc@{}}
    \toprule
    \textbf{Data} & \textbf{Model} & \textbf{Camera} &
    \textbf{Robot} & \textbf{Language} & \textbf{Light} &
    \textbf{Background} & \textbf{Noise} & \textbf{Layout} &
    \textbf{Average} \\
    \midrule
\multirow{5}{*}{\textbf{25\%}} & Baseline & \underline{37.7} & \underline{32.4} & 75.2 & 82.8 & \underline{79.1} & 52.1 & \underline{57.6} & 57.5 \\
 & WM-Align & 35.6 & 29.9 & \underline{78.1} & \underline{86.6} & 77.3 & \textbf{60.1} & 56.5 & \underline{58.6} \\
 & WM-Query & 26.4 & 27.0 & 69.6 & 75.8 & 63.0 & 41.5 & 46.3 & 48.1 \\
 & WM-MoT & \textbf{41.7} & \textbf{37.5} & \textbf{80.9} & \textbf{87.6} & \textbf{84.3} & \underline{59.8} & \textbf{59.0} & \textbf{62.4} \\
 & WM-Ext & 37.4 & 28.9 & 76.3 & 81.8 & 73.8 & 56.2 & 55.5 & 56.8 \\
\midrule
\multirow{5}{*}{\textbf{50\%}} & Baseline & 46.2 & \underline{41.9} & 84.0 & 87.9 & 85.9 & 53.6 & 67.0 & 64.7 \\
 & WM-Align & \underline{54.0} & 38.6 & 84.2 & 88.6 & \textbf{89.0} & 65.9 & 66.9 & 67.8 \\
 & WM-Query & 42.4 & 39.2 & \textbf{85.2} & 92.3 & 87.6 & 61.5 & 63.0 & 65.2 \\
 & WM-MoT & \textbf{57.5} & \textbf{47.7} & \underline{85.1} & \textbf{93.5} & \underline{88.9} & \underline{71.6} & \underline{68.8} & \textbf{71.7} \\
 & WM-Ext & 46.6 & 39.4 & 83.1 & \underline{93.4} & 88.5 & \textbf{72.3} & \textbf{69.2} & \underline{68.4} \\
\midrule
\multirow{5}{*}{\textbf{100\%}} & Baseline & 48.2 & \underline{51.4} & \underline{91.4} & 96.0 & 92.9 & 65.3 & 71.3 & 71.8 \\
 & WM-Align & \underline{58.0} & 49.7 & 89.5 & 94.6 & 93.5 & 71.8 & 73.9 & 74.1 \\
 & WM-Query & 54.4 & 50.6 & 88.5 & 95.1 & \textbf{96.3} & 69.5 & 69.5 & 72.9 \\
 & WM-MoT & \textbf{59.7} & \textbf{53.5} & \textbf{91.6} & \textbf{97.6} & \underline{95.9} & \underline{72.3} & \underline{75.2} & \textbf{76.2} \\
 & WM-Ext & 55.7 & 46.8 & 91.2 & \underline{96.8} & 95.5 & \textbf{74.2} & \textbf{75.9} & \underline{74.7} \\
    \bottomrule
  \end{tabular*}
\end{table*}

\subsection{Detailed RoboTwin Results}
\label{app:robotwin_details}

Table~\ref{tab:app-robotwin-per-task} reports the task-level success rates
for the baseline and world-modeling variants underlying
Table~\ref{tab:robotwin-c2c-c2r}.

\begin{table*}[htbp]
    \centering
    \caption{Per-task success rates (\%) for the action-only baseline and
    world-modeling variants on RoboTwin 2.0. Group and overall averages
    correspond to the respective rows in Table~\ref{tab:robotwin-c2c-c2r}.}
    \label{tab:app-robotwin-per-task}
    \scriptsize
    \setlength{\tabcolsep}{2.5pt}
    \renewcommand{\arraystretch}{0.92}
    \resizebox{\textwidth}{!}{%
    \begin{tabular}{@{}lcccccccccc@{}}
        \toprule
        \multirow{2}{*}{\textbf{Task}}
        & \multicolumn{5}{c}{\textbf{Clean2Clean}}
        & \multicolumn{5}{c}{\textbf{Clean2Random}} \\
        \cmidrule(lr){2-6}\cmidrule(l){7-11}
        & Baseline & WM-Align & WM-Query & WM-MoT & WM-Ext
        & Baseline & WM-Align & WM-Query & WM-MoT & WM-Ext \\
        \midrule
        \multicolumn{11}{@{}l}{\textit{Short-horizon tasks}} \\
        adjust\_bottle & 100 & 100 & 100 & 100 & 99 & 8 & 13 & 3 & 4 & 3 \\
        beat\_block\_hammer & 58 & 87 & 84 & 67 & 69 & 1 & 6 & 9 & 2 & 2 \\
        click\_alarmclock & 9 & 3 & 2 & 2 & 2 & 71 & 63 & 47 & 61 & 55 \\
        click\_bell & 3 & 19 & 8 & 33 & 11 & 64 & 61 & 41 & 54 & 38 \\
        grab\_roller & 99 & 100 & 97 & 100 & 100 & 65 & 56 & 49 & 25 & 43 \\
        lift\_pot & 80 & 90 & 86 & 83 & 79 & 5 & 2 & 5 & 6 & 2 \\
        move\_can\_pot & 22 & 48 & 40 & 39 & 46 & 1 & 6 & 1 & 1 & 2 \\
        move\_pillbottle\_pad & 39 & 75 & 66 & 62 & 67 & 0 & 3 & 0 & 3 & 1 \\
        move\_playingcard\_away & 81 & 85 & 86 & 85 & 86 & 3 & 5 & 5 & 1 & 3 \\
        move\_stapler\_pad & 20 & 24 & 20 & 27 & 20 & 0 & 0 & 0 & 0 & 0 \\
        pick\_diverse\_bottles & 28 & 49 & 49 & 35 & 38 & 9 & 12 & 11 & 12 & 9 \\
        pick\_dual\_bottles & 50 & 68 & 69 & 54 & 61 & 8 & 18 & 11 & 20 & 24 \\
        place\_a2b\_left & 39 & 39 & 41 & 35 & 30 & 5 & 8 & 8 & 9 & 5 \\
        place\_a2b\_right & 33 & 32 & 35 & 36 & 30 & 5 & 12 & 3 & 5 & 3 \\
        place\_bread\_skillet & 69 & 73 & 76 & 81 & 75 & 4 & 7 & 3 & 9 & 3 \\
        place\_container\_plate & 95 & 99 & 98 & 97 & 98 & 25 & 30 & 31 & 34 & 26 \\
        place\_empty\_cup & 96 & 98 & 96 & 97 & 96 & 12 & 18 & 19 & 14 & 10 \\
        place\_fan & 36 & 54 & 52 & 54 & 56 & 1 & 1 & 1 & 3 & 1 \\
        place\_mouse\_pad & 27 & 35 & 44 & 34 & 27 & 0 & 1 & 2 & 1 & 0 \\
        place\_object\_scale & 38 & 45 & 49 & 48 & 51 & 1 & 2 & 2 & 2 & 1 \\
        place\_object\_stand & 72 & 80 & 79 & 76 & 79 & 8 & 14 & 7 & 14 & 5 \\
        place\_phone\_stand & 25 & 40 & 27 & 36 & 22 & 1 & 3 & 2 & 3 & 2 \\
        place\_shoe & 80 & 75 & 78 & 80 & 77 & 9 & 12 & 9 & 9 & 6 \\
        press\_stapler & 94 & 95 & 95 & 91 & 94 & 57 & 60 & 47 & 49 & 55 \\
        rotate\_qrcode & 62 & 74 & 80 & 81 & 76 & 2 & 2 & 2 & 1 & 1 \\
        scan\_object & 64 & 73 & 71 & 77 & 77 & 6 & 6 & 2 & 5 & 4 \\
        stamp\_seal & 42 & 47 & 42 & 45 & 46 & 1 & 1 & 3 & 0 & 0 \\
        turn\_switch & 35 & 30 & 41 & 34 & 37 & 3 & 6 & 9 & 6 & 1 \\
        \rowcolor{black!8}
        \textit{Group average} & 53.4 & 62.0 & 61.1 & 60.3 & 58.9 & 13.4 & 15.3 & 11.9 & 12.6 & 10.9 \\
        \addlinespace[2pt]
        \multicolumn{11}{@{}l}{\textit{Medium-horizon tasks}} \\
        dump\_bin\_bigbin & 88 & 96 & 97 & 89 & 96 & 51 & 49 & 32 & 63 & 37 \\
        handover\_block & 34 & 34 & 51 & 51 & 27 & 0 & 0 & 0 & 0 & 0 \\
        handover\_mic & 67 & 85 & 89 & 95 & 89 & 1 & 4 & 2 & 3 & 2 \\
        open\_laptop & 82 & 81 & 90 & 84 & 88 & 8 & 7 & 12 & 10 & 8 \\
        place\_bread\_basket & 66 & 71 & 74 & 68 & 68 & 7 & 20 & 18 & 10 & 10 \\
        place\_burger\_fries & 91 & 94 & 98 & 96 & 96 & 8 & 17 & 4 & 18 & 3 \\
        place\_can\_basket & 19 & 33 & 46 & 41 & 39 & 0 & 4 & 4 & 6 & 3 \\
        place\_cans\_plasticbox & 66 & 83 & 93 & 79 & 92 & 2 & 1 & 6 & 8 & 5 \\
        place\_dual\_shoes & 48 & 51 & 56 & 49 & 50 & 2 & 7 & 6 & 6 & 2 \\
        place\_object\_basket & 75 & 76 & 83 & 78 & 78 & 5 & 8 & 5 & 17 & 1 \\
        put\_object\_cabinet & 23 & 40 & 35 & 37 & 37 & 4 & 6 & 5 & 5 & 3 \\
        shake\_bottle & 95 & 99 & 100 & 97 & 100 & 56 & 52 & 58 & 55 & 71 \\
        shake\_bottle\_horizontally & 91 & 98 & 99 & 99 & 99 & 56 & 58 & 54 & 60 & 68 \\
        \rowcolor{black!8}
        \textit{Group average} & 65.0 & 72.4 & 77.8 & 74.1 & 73.8 & 15.4 & 17.9 & 15.8 & 20.1 & 16.4 \\
        \addlinespace[2pt]
        \multicolumn{11}{@{}l}{\textit{Long-horizon tasks}} \\
        blocks\_ranking\_rgb & 41 & 63 & 69 & 74 & 59 & 0 & 2 & 5 & 6 & 2 \\
        blocks\_ranking\_size & 22 & 30 & 34 & 34 & 22 & 1 & 1 & 2 & 2 & 1 \\
        hanging\_mug & 17 & 14 & 39 & 17 & 20 & 0 & 0 & 0 & 0 & 1 \\
        open\_microwave & 10 & 16 & 24 & 26 & 18 & 39 & 31 & 24 & 21 & 29 \\
        put\_bottles\_dustbin & 15 & 54 & 57 & 55 & 46 & 5 & 14 & 21 & 12 & 12 \\
        stack\_blocks\_three & 7 & 31 & 37 & 32 & 24 & 0 & 0 & 0 & 1 & 0 \\
        stack\_blocks\_two & 59 & 81 & 85 & 82 & 79 & 3 & 4 & 5 & 4 & 5 \\
        stack\_bowls\_three & 35 & 64 & 74 & 69 & 64 & 4 & 12 & 11 & 17 & 6 \\
        stack\_bowls\_two & 70 & 86 & 94 & 87 & 93 & 7 & 16 & 25 & 25 & 15 \\
        \rowcolor{black!8}
        \textit{Group average} & 30.7 & 48.8 & 57.0 & 52.9 & 47.2 & 6.6 & 8.9 & 10.3 & 9.8 & 7.9 \\
        \midrule
        \textbf{Overall average}
        & \textbf{52.3} & \textbf{62.3} & \textbf{64.7} & \textbf{62.6} & \textbf{60.7}
        & \textbf{12.7} & \textbf{14.8} & \textbf{12.6} & \textbf{14.0} & \textbf{11.8} \\
        \bottomrule
    \end{tabular}%
    }
\end{table*}

\subsection{Detailed \texorpdfstring{$\pi_{0.5}$}{pi0.5} RoboTwin Results}
\label{app:pi05_robotwin_details}

Table~\ref{tab:app-pi05-robotwin-per-task} reports the task-level success
rates underlying Table~\ref{tab:pi05-robotwin-c2c-c2r}.

\begin{table*}[htbp]
    \centering
    \caption{Per-task success rates (\%) for the $\pi_{0.5}$ experiments on
    RoboTwin 2.0 under Clean2Clean and Clean2Random evaluation. Group and
    overall averages correspond to Table~\ref{tab:pi05-robotwin-c2c-c2r}.}
    \label{tab:app-pi05-robotwin-per-task}
    \scriptsize
    \setlength{\tabcolsep}{3.0pt}
    \renewcommand{\arraystretch}{0.86}
    \resizebox{\textwidth}{!}{%
    \begin{tabular}{@{}lcccccccccc@{}}
        \toprule
        \multirow{2}{*}{\textbf{Task}}
        & \multicolumn{5}{c}{\textbf{Clean2Clean}}
        & \multicolumn{5}{c}{\textbf{Clean2Random}} \\
        \cmidrule(lr){2-6}\cmidrule(l){7-11}
        & $\pi_{0.5}$ & WM-Align & WM-Query & WM-MoT & WM-Ext
        & $\pi_{0.5}$ & WM-Align & WM-Query & WM-MoT & WM-Ext \\
        \midrule
        \multicolumn{11}{@{}l}{\textit{Short-horizon tasks}} \\
        adjust\_bottle & 100 & 100 & 100 & 100 & 100 & 90 & 90 & 96 & 93 & 97 \\
        beat\_block\_hammer & 83 & 89 & 92 & 92 & 90 & 29 & 21 & 25 & 37 & 27 \\
        click\_alarmclock & 63 & 50 & 49 & 56 & 58 & 57 & 50 & 56 & 52 & 60 \\
        click\_bell & 59 & 24 & 44 & 46 & 57 & 44 & 25 & 34 & 46 & 48 \\
        grab\_roller & 99 & 100 & 99 & 100 & 100 & 98 & 96 & 96 & 98 & 100 \\
        lift\_pot & 94 & 96 & 95 & 95 & 97 & 39 & 22 & 35 & 48 & 42 \\
        move\_can\_pot & 70 & 71 & 75 & 65 & 81 & 29 & 19 & 21 & 32 & 30 \\
        move\_pillbottle\_pad & 42 & 26 & 47 & 44 & 54 & 31 & 23 & 28 & 36 & 38 \\
        move\_playingcard\_away & 81 & 70 & 71 & 85 & 81 & 55 & 61 & 60 & 65 & 59 \\
        move\_stapler\_pad & 25 & 23 & 21 & 22 & 24 & 19 & 19 & 21 & 17 & 20 \\
        pick\_diverse\_bottles & 71 & 64 & 71 & 69 & 72 & 44 & 42 & 47 & 47 & 49 \\
        pick\_dual\_bottles & 77 & 86 & 78 & 76 & 83 & 58 & 62 & 55 & 66 & 62 \\
        place\_a2b\_left & 67 & 64 & 66 & 73 & 75 & 30 & 44 & 40 & 43 & 44 \\
        place\_a2b\_right & 64 & 59 & 69 & 70 & 71 & 34 & 19 & 35 & 38 & 36 \\
        place\_bread\_skillet & 60 & 52 & 64 & 76 & 63 & 36 & 29 & 31 & 39 & 37 \\
        place\_container\_plate & 95 & 97 & 96 & 97 & 95 & 79 & 74 & 80 & 83 & 77 \\
        place\_empty\_cup & 96 & 92 & 94 & 93 & 96 & 81 & 73 & 81 & 77 & 79 \\
        place\_fan & 60 & 56 & 60 & 69 & 67 & 36 & 30 & 33 & 46 & 38 \\
        place\_mouse\_pad & 29 & 22 & 25 & 28 & 31 & 27 & 20 & 20 & 29 & 25 \\
        place\_object\_scale & 68 & 61 & 71 & 71 & 74 & 44 & 33 & 46 & 46 & 47 \\
        place\_object\_stand & 82 & 77 & 81 & 85 & 85 & 55 & 44 & 64 & 63 & 64 \\
        place\_phone\_stand & 50 & 38 & 52 & 59 & 46 & 41 & 34 & 40 & 43 & 39 \\
        place\_shoe & 69 & 67 & 74 & 68 & 71 & 65 & 54 & 59 & 62 & 61 \\
        press\_stapler & 83 & 89 & 85 & 81 & 89 & 59 & 52 & 64 & 71 & 65 \\
        rotate\_qrcode & 87 & 89 & 84 & 89 & 92 & 32 & 26 & 32 & 32 & 33 \\
        scan\_object & 29 & 31 & 29 & 40 & 30 & 20 & 14 & 25 & 26 & 20 \\
        stamp\_seal & 48 & 54 & 52 & 49 & 54 & 23 & 21 & 26 & 31 & 24 \\
        turn\_switch & 50 & 48 & 55 & 51 & 66 & 21 & 19 & 21 & 23 & 25 \\
        \rowcolor{black!8}
        \textit{Group average} & 67.9 & 64.1 & 67.8 & 69.6 & 71.5 & 45.6 & 39.9 & 45.4 & 49.6 & 48.1 \\
        \addlinespace[2pt]
        \multicolumn{11}{@{}l}{\textit{Medium-horizon tasks}} \\
        dump\_bin\_bigbin & 90 & 88 & 89 & 95 & 93 & 73 & 74 & 66 & 75 & 74 \\
        handover\_block & 55 & 41 & 63 & 49 & 49 & 23 & 11 & 19 & 17 & 20 \\
        handover\_mic & 97 & 95 & 95 & 94 & 95 & 12 & 20 & 11 & 15 & 14 \\
        open\_laptop & 91 & 89 & 91 & 88 & 95 & 78 & 51 & 72 & 73 & 74 \\
        place\_bread\_basket & 65 & 47 & 49 & 68 & 60 & 55 & 41 & 56 & 59 & 53 \\
        place\_burger\_fries & 69 & 68 & 65 & 81 & 73 & 77 & 72 & 68 & 68 & 76 \\
        place\_can\_basket & 55 & 68 & 67 & 64 & 67 & 15 & 13 & 13 & 18 & 13 \\
        place\_cans\_plasticbox & 67 & 35 & 57 & 75 & 66 & 61 & 49 & 50 & 52 & 55 \\
        place\_dual\_shoes & 53 & 38 & 59 & 67 & 59 & 38 & 36 & 42 & 33 & 40 \\
        place\_object\_basket & 72 & 75 & 74 & 80 & 81 & 41 & 25 & 39 & 42 & 39 \\
        put\_object\_cabinet & 29 & 28 & 32 & 38 & 37 & 20 & 26 & 23 & 18 & 26 \\
        shake\_bottle & 100 & 99 & 100 & 100 & 100 & 100 & 97 & 100 & 100 & 97 \\
        shake\_bottle\_horizontally & 99 & 97 & 100 & 100 & 99 & 100 & 98 & 100 & 100 & 100 \\
        \rowcolor{black!8}
        \textit{Group average} & 72.5 & 66.8 & 72.4 & 76.8 & 74.9 & 53.3 & 47.2 & 50.7 & 51.5 & 52.4 \\
        \addlinespace[2pt]
        \multicolumn{11}{@{}l}{\textit{Long-horizon tasks}} \\
        blocks\_ranking\_rgb & 72 & 68 & 73 & 74 & 73 & 44 & 40 & 47 & 56 & 51 \\
        blocks\_ranking\_size & 42 & 38 & 45 & 51 & 47 & 16 & 10 & 17 & 24 & 22 \\
        hanging\_mug & 14 & 13 & 21 & 19 & 19 & 7 & 13 & 13 & 17 & 14 \\
        open\_microwave & 82 & 58 & 81 & 78 & 72 & 30 & 38 & 53 & 38 & 45 \\
        put\_bottles\_dustbin & 65 & 58 & 57 & 66 & 68 & 47 & 45 & 44 & 56 & 49 \\
        stack\_blocks\_three & 73 & 43 & 64 & 71 & 73 & 25 & 10 & 18 & 36 & 25 \\
        stack\_blocks\_two & 90 & 70 & 84 & 87 & 89 & 45 & 29 & 44 & 55 & 49 \\
        stack\_bowls\_three & 80 & 75 & 72 & 78 & 76 & 54 & 56 & 65 & 68 & 61 \\
        stack\_bowls\_two & 96 & 95 & 97 & 93 & 94 & 83 & 82 & 80 & 75 & 81 \\
        \rowcolor{black!8}
        \textit{Group average} & 68.2 & 57.6 & 66.0 & 68.6 & 67.9 & 39.0 & 35.9 & 42.3 & 47.2 & 44.1 \\
        \midrule
        \textbf{Overall average}
        & \textbf{69.1} & \textbf{63.6} & \textbf{68.7} & \textbf{71.3} & \textbf{71.7}
        & \textbf{46.4} & \textbf{41.0} & \textbf{46.2} & \textbf{49.7} & \textbf{48.5} \\
        \bottomrule
    \end{tabular}%
    }
\end{table*}

\subsection{Detailed DOMINO Results}
\label{app:domino_details}

Table~\ref{tab:app-domino-per-task} reports the task-level success rate
and manipulation score underlying Table~\ref{tab:dynamic-environments}.

\begin{table*}[htbp]
    \centering
    \caption{Per-task performance on DOMINO Level~1 under the clean dynamic
    setting, reported as success rate (SR) and manipulation score (MS).}
    \label{tab:app-domino-per-task}
    \scriptsize
    \setlength{\tabcolsep}{3.0pt}
    \renewcommand{\arraystretch}{1.00}
    \resizebox{\textwidth}{!}{%
    \begin{tabular}{@{}l*{5}{cc}@{}}
        \toprule
        \multirow{2}{*}{\textbf{Task}}
        & \multicolumn{2}{c}{\textbf{Baseline}}
        & \multicolumn{2}{c}{\textbf{WM-Align}}
        & \multicolumn{2}{c}{\textbf{WM-Query}}
        & \multicolumn{2}{c}{\textbf{WM-MoT}}
        & \multicolumn{2}{c}{\textbf{WM-Ext}} \\
        \cmidrule(lr){2-3}\cmidrule(lr){4-5}\cmidrule(lr){6-7}
        \cmidrule(lr){8-9}\cmidrule(l){10-11}
        & SR & MS & SR & MS & SR & MS & SR & MS & SR & MS \\
        \midrule
        adjust\_bottle & 63 & 77.60 & 59 & 74.60 & 51 & 71.40 & 69 & 80.80 & 63 & 76.50 \\
        beat\_block\_hammer & 9 & 21.70 & 7 & 18.00 & 4 & 16.60 & 6 & 18.40 & 8 & 21.70 \\
        click\_alarmclock & 12 & 18.00 & 11 & 17.00 & 11 & 16.80 & 14 & 18.60 & 16 & 21.30 \\
        click\_bell & 2 & 11.90 & 0 & 11.20 & 1 & 10.70 & 0 & 9.00 & 0 & 11.00 \\
        dump\_bin\_bigbin & 19 & 35.00 & 28 & 38.80 & 12 & 33.00 & 29 & 43.00 & 34 & 46.60 \\
        grab\_roller & 22 & 47.30 & 30 & 54.30 & 29 & 53.30 & 27 & 53.50 & 29 & 54.30 \\
        handover\_block & 3 & 48.70 & 1 & 45.90 & 6 & 59.00 & 10 & 54.50 & 4 & 47.60 \\
        handover\_mic & 5 & 32.30 & 8 & 35.20 & 8 & 36.50 & 15 & 40.60 & 13 & 38.50 \\
        hanging\_mug & 7 & 37.00 & 6 & 36.80 & 5 & 38.40 & 7 & 37.30 & 6 & 36.50 \\
        move\_can\_pot & 16 & 44.80 & 22 & 45.90 & 19 & 52.10 & 23 & 50.20 & 23 & 51.00 \\
        move\_pillbottle\_pad & 5 & 16.60 & 4 & 13.80 & 4 & 14.60 & 6 & 16.10 & 6 & 17.20 \\
        move\_playingcard\_away & 1 & 24.10 & 3 & 26.40 & 8 & 28.80 & 7 & 29.40 & 7 & 28.30 \\
        move\_stapler\_pad & 1 & 14.30 & 0 & 10.90 & 0 & 11.90 & 1 & 14.10 & 2 & 14.50 \\
        place\_a2b\_left & 1 & 25.30 & 0 & 25.50 & 2 & 28.10 & 4 & 25.20 & 3 & 27.70 \\
        place\_a2b\_right & 0 & 23.80 & 2 & 25.50 & 1 & 25.40 & 2 & 28.80 & 2 & 26.90 \\
        place\_bread\_basket & 6 & 18.60 & 6 & 18.40 & 20 & 30.80 & 20 & 29.80 & 21 & 30.10 \\
        place\_bread\_skillet & 15 & 30.90 & 14 & 30.50 & 14 & 30.40 & 18 & 31.60 & 17 & 33.10 \\
        place\_can\_basket & 8 & 33.90 & 10 & 37.30 & 12 & 41.70 & 10 & 37.10 & 10 & 37.90 \\
        place\_container\_plate & 21 & 27.60 & 16 & 23.40 & 25 & 31.30 & 32 & 37.80 & 26 & 33.10 \\
        place\_empty\_cup & 5 & 16.50 & 9 & 22.10 & 4 & 16.10 & 7 & 18.60 & 6 & 18.70 \\
        place\_fan & 1 & 8.30 & 1 & 9.90 & 3 & 9.80 & 4 & 11.30 & 5 & 11.70 \\
        place\_mouse\_pad & 1 & 15.00 & 0 & 14.10 & 3 & 15.70 & 5 & 18.50 & 3 & 16.30 \\
        place\_object\_basket & 4 & 25.10 & 11 & 33.00 & 15 & 33.40 & 15 & 35.60 & 18 & 36.70 \\
        place\_object\_scale & 0 & 12.10 & 0 & 11.80 & 0 & 11.80 & 0 & 11.70 & 2 & 13.10 \\
        place\_object\_stand & 2 & 11.60 & 0 & 10.40 & 4 & 13.60 & 3 & 12.60 & 4 & 14.20 \\
        place\_phone\_stand & 0 & 17.10 & 0 & 16.70 & 0 & 17.00 & 2 & 17.20 & 4 & 18.60 \\
        place\_shoe & 10 & 23.80 & 15 & 27.90 & 18 & 30.20 & 21 & 32.90 & 23 & 35.50 \\
        press\_stapler & 10 & 21.90 & 13 & 25.10 & 15 & 23.20 & 14 & 23.60 & 12 & 23.70 \\
        put\_bottles\_dustbin & 23 & 43.20 & 26 & 47.20 & 36 & 55.70 & 39 & 54.20 & 31 & 51.00 \\
        put\_object\_cabinet & 15 & 53.70 & 14 & 51.60 & 24 & 58.10 & 22 & 57.60 & 23 & 57.50 \\
        rotate\_qrcode & 4 & 25.00 & 8 & 28.10 & 9 & 29.80 & 9 & 29.40 & 12 & 31.30 \\
        scan\_object & 2 & 28.80 & 5 & 31.50 & 4 & 29.20 & 5 & 34.20 & 4 & 31.20 \\
        shake\_bottle & 32 & 49.70 & 44 & 55.90 & 49 & 61.70 & 39 & 53.50 & 39 & 56.00 \\
        shake\_bottle\_horizontally & 44 & 60.40 & 40 & 59.00 & 49 & 64.00 & 53 & 66.10 & 45 & 62.90 \\
        stamp\_seal & 3 & 17.90 & 5 & 19.50 & 8 & 21.40 & 6 & 20.40 & 8 & 22.20 \\
        \midrule
        \textbf{Average} & \textbf{10.6} & \textbf{29.1}
        & \textbf{11.9} & \textbf{30.1}
        & \textbf{13.5} & \textbf{32.0}
        & \textbf{15.5} & \textbf{32.9}
        & \textbf{15.1} & \textbf{33.0} \\
        \bottomrule
    \end{tabular}%
    }
\end{table*}

\clearpage
\subsection{Detailed Real-Robot Results}
\label{app:real_robot_details}

Table~\ref{tab:app-real-robot-per-task} reports the complete per-task
success rates across the static and dynamic tasks evaluated on the physical
Franka Research 3, corresponding to the real-world findings in
Section~\ref{subsec:real_world_experiments}. Each task is evaluated over
40 trials per method.

\begin{table*}[htbp]
  \centering
  \caption{Real-world manipulation success rates (\%) on the physical Franka
  Research 3. Each task is evaluated over 40 trials. \textbf{Bold} and
  \underline{underline} denote the best and second-best results, respectively.}
  \label{tab:app-real-robot-per-task}
  \small
  \setlength{\tabcolsep}{4.5pt}
  \renewcommand{\arraystretch}{1.12}
  \begin{tabular*}{\textwidth}{@{\extracolsep{\fill}}lccccccc@{}}
    \toprule
    \multirow{2}{*}{\textbf{Method}}
    & \multicolumn{3}{c}{\textbf{Static Tasks}}
    & \multicolumn{3}{c}{\textbf{Dynamic Tasks}}
    & \multirow{2}{*}{\textbf{Overall}} \\
    \cmidrule(lr){2-4}\cmidrule(lr){5-7}
    & \shortstack{Box Cover\\Block}
    & \shortstack{Place Corn\\on Rack}
    & \textbf{Avg.}
    & \shortstack{Pick\\Tomato}
    & \shortstack{Place Tomato\\on Plate}
    & \textbf{Avg.}
    & \\
    \midrule
    Baseline & 62.5 & 60.0 & 61.3 & 37.5 & 20.0 & 28.8 & 45.0 \\
    \addlinespace[2pt]
    WM-Align & 77.5 & \textbf{85.0} & \underline{81.3} & 47.5 & 25.0 & 36.3 & 58.8 \\
    WM-Query & 67.5 & 65.0 & 66.3 & 57.5 & 35.0 & 46.3 & 56.3 \\
    WM-MoT & \underline{87.5} & \underline{77.5} & \textbf{82.5}
    & \textbf{75.0} & \textbf{55.0} & \textbf{65.0} & \textbf{73.8} \\
    WM-Ext & \textbf{90.0} & 67.5 & 78.8
    & \underline{70.0} & \underline{42.5} & \underline{56.3}
    & \underline{67.5} \\
    \bottomrule
  \end{tabular*}
\end{table*}

\clearpage
\subsection{Qualitative Real-Robot Rollouts}
\label{app:real_robot_rollouts}

\begin{figure}[p]
    \centering
    \includegraphics[width=\textwidth]{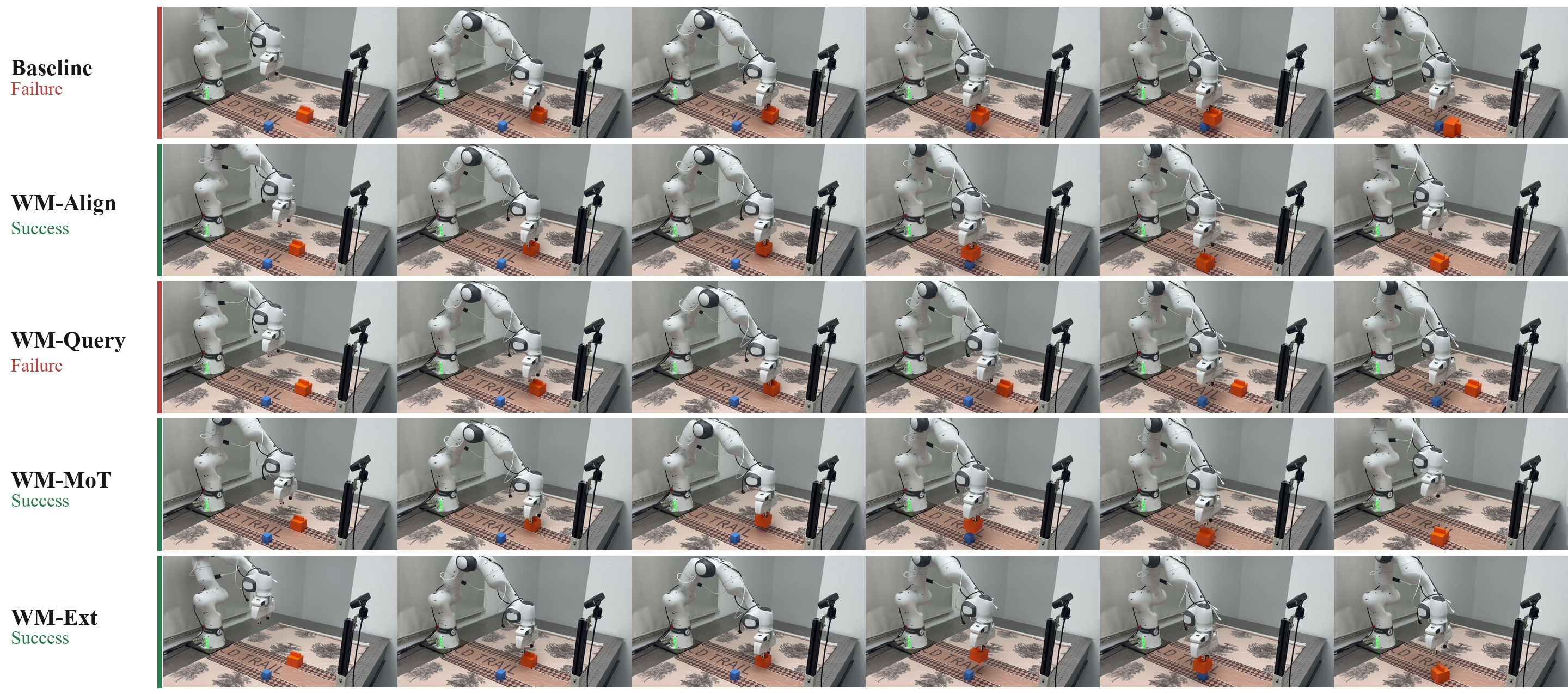}
    \caption{\textbf{Qualitative real-robot rollouts for \emph{Box Cover Block}.}
    Each row shows one selected rollout, with its method and outcome indicated
    on the left.}
    \label{fig:real_robot_rollout_box}
\end{figure}

\begin{figure}[p]
    \centering
    \includegraphics[width=\textwidth]{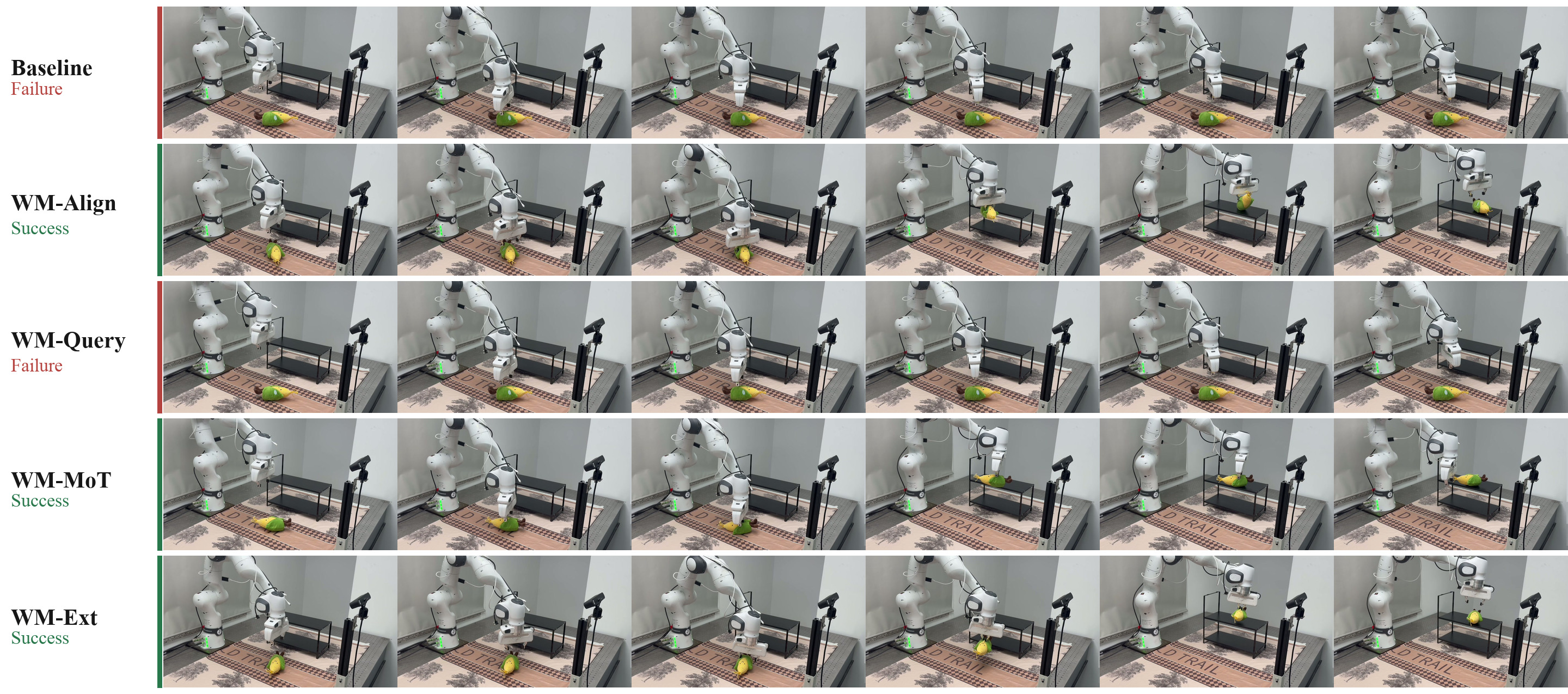}
    \caption{\textbf{Qualitative real-robot rollouts for \emph{Place Corn on Rack}.}
    Each row shows one selected rollout, with its method and outcome indicated
    on the left.}
    \label{fig:real_robot_rollout_corn}
\end{figure}

\begin{figure}[p]
    \centering
    \includegraphics[width=\textwidth]{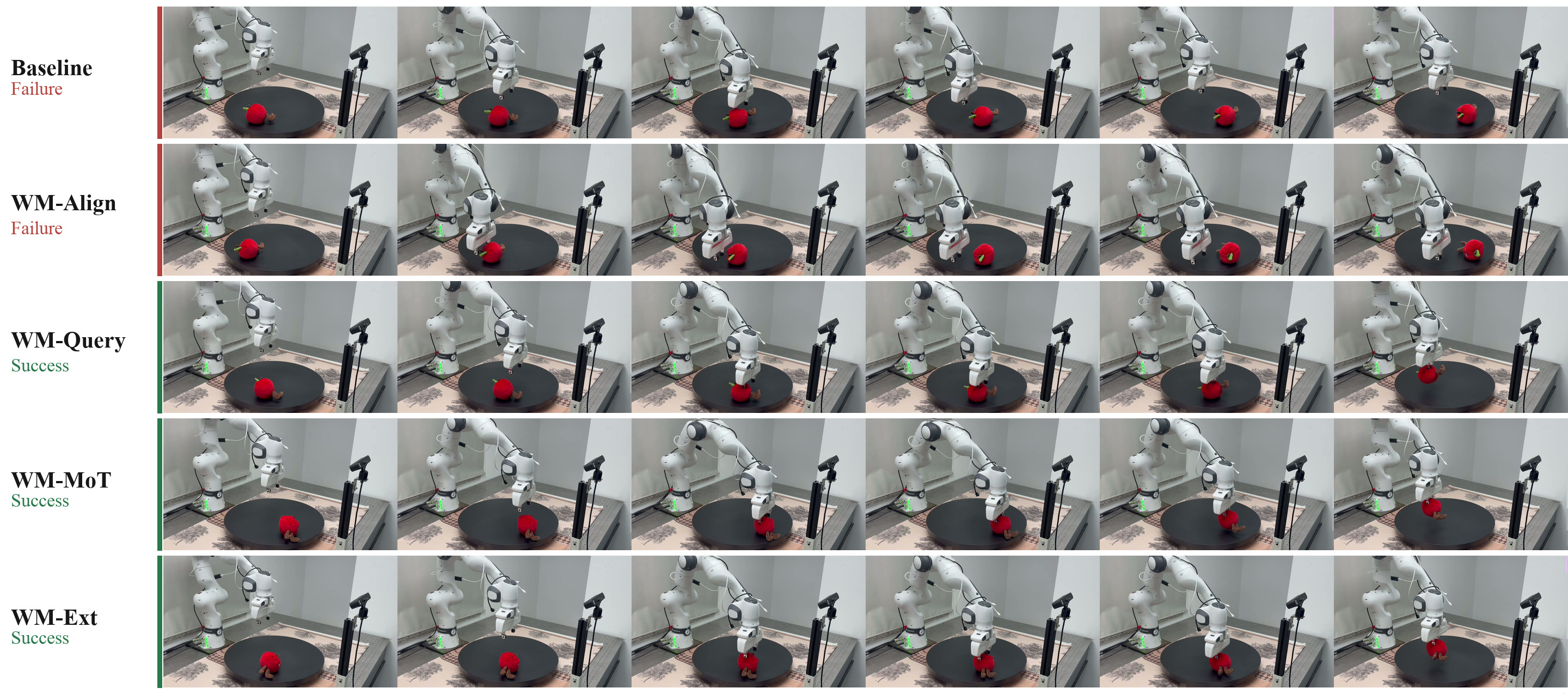}
    \caption{\textbf{Qualitative real-robot rollouts for \emph{Pick Tomato}.}
    Each row shows one selected rollout, with its method and outcome indicated
    on the left.}
    \label{fig:real_robot_rollout_pick}
\end{figure}

\begin{figure}[p]
    \centering
    \includegraphics[width=\textwidth]{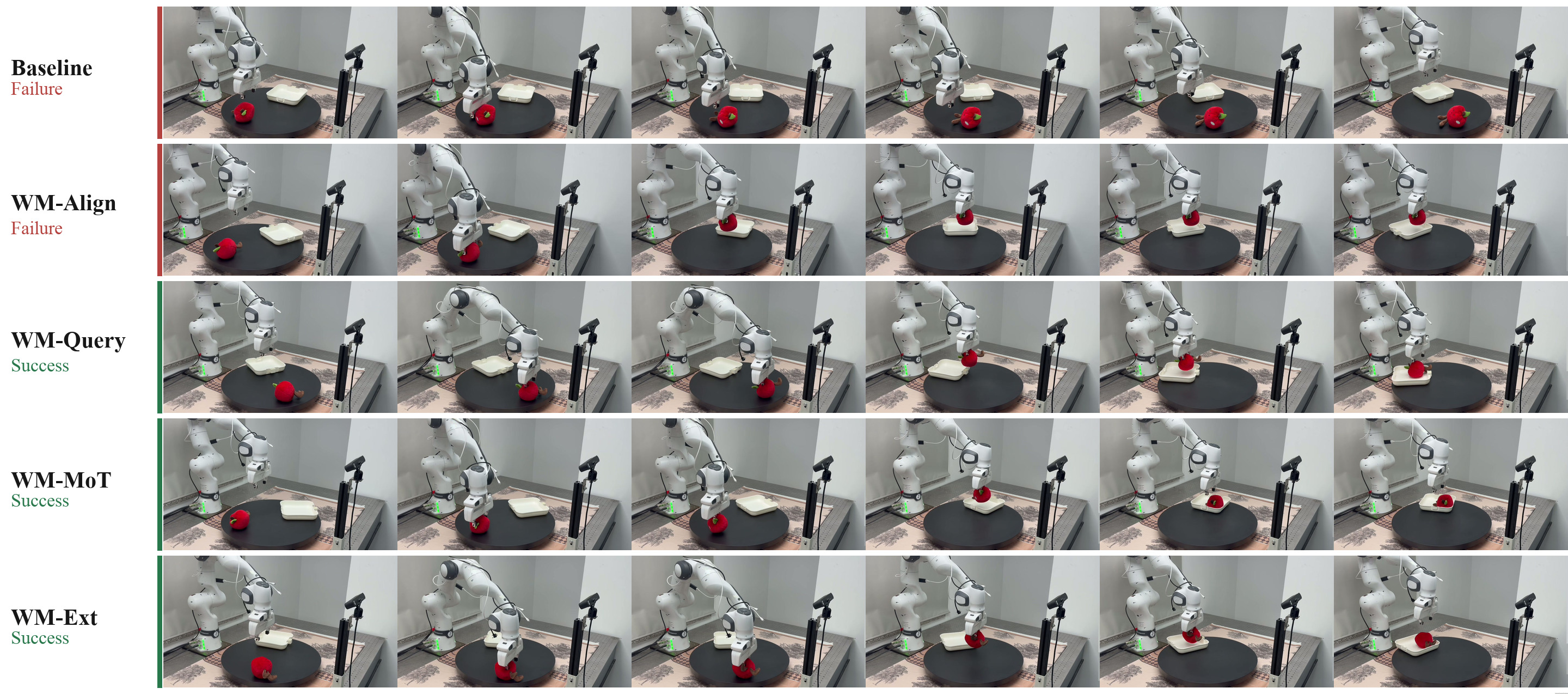}
    \caption{\textbf{Qualitative real-robot rollouts for \emph{Place Tomato on Plate}.}
    Each row shows one selected rollout, with its method and outcome indicated
    on the left.}
    \label{fig:real_robot_rollout_plate}
\end{figure}

Figures~\ref{fig:real_robot_rollout_box}--\ref{fig:real_robot_rollout_plate}
show representative physical-robot rollouts for all four tasks. Each row contains six uniformly sampled frames from one trajectory.

\clearpage

\end{document}